\documentclass{article}
\usepackage[final,nonatbib]{neurips_2020}

\usepackage[utf8]{inputenc} %
\usepackage[T1]{fontenc}    %

\usepackage{algorithm}
\usepackage[noend]{algpseudocode}
\usepackage{algorithmicx}

\usepackage{microtype}
\usepackage{lipsum}
\usepackage{graphicx}
\usepackage{subcaption}
\usepackage{caption}
\usepackage{booktabs} %
\usepackage[hidelinks]{hyperref}       %
\usepackage{url}            %
\usepackage{booktabs}       %
\usepackage{amsfonts}       %
\usepackage{nicefrac}       %
\usepackage{microtype}      %
\usepackage{comment}
\usepackage{enumitem}
\usepackage{wrapfig,tikz}
\usepackage{amsmath,longtable,fancyhdr,booktabs}
\usepackage{adjustbox, bigstrut, tabularx, multirow, makecell, diagbox}
\usepackage{graphicx,float}
\usepackage{amssymb,xcolor,amsthm}
\usepackage{stackrel}
\usepackage{color}
\usepackage{colortbl}
\usepackage{theoremref}
\usepackage{stmaryrd}
\usepackage[capitalise]{cleveref}

\newcommand*{\tran}{^{\mkern-1.5mu\mathsf{T}}}
\newcommand{\mbf}[1]{\mathbf{#1}}

\newcommand{\mbb}[1]{\mathbb{#1}}
\newcolumntype{C}[1]{>{\centering\arraybackslash}m{#1}}
\newcolumntype{P}[1]{>{\centering\arraybackslash}p{#1}}
\newcommand{\ud}{\mathrm{d}}

\newcommand{\mcal}{\mathcal}
\newcommand{\norm}[1]{\left\lVert#1\right\rVert}

\usepackage{mdframed}
\newtheorem{recipe}{Technique}

\newtheorem{proposition}{Proposition}

\newtheorem{innercustomprop}{Proposition}
\newenvironment{customprop}[1]
{\renewcommand\theinnercustomprop{#1}\innercustomprop}
{\endinnercustomprop}

\newcommand{\be}{\begin{equation}}
\newcommand{\ee}{\end{equation}}
\definecolor{Gray}{gray}{0.85}
\definecolor{LightCyan}{rgb}{0.88,1,1}

\makeatletter
\usepackage{xspace} 
\def\@onedot{\ifx\@let@token.\else.\null\fi\xspace}
\DeclareRobustCommand\onedot{\futurelet\@let@token\@onedot}

\newcommand{\bfx}{\mathbf{x}}

\newcommand{\bfz}{\mathbf{z}}
\newcommand{\bfI}{\mathbf{I}}

\newcommand{\bfzero}{\mathbf{0}}

\newcommand{\bftheta}{{\boldsymbol{\theta}}}

\newcommand{\bfphi}{{\boldsymbol{\phi}}}

\newcommand{\bfs}{\mathbf{s}}

\def\eg{\emph{e.g}\onedot}

\def\ie{\emph{i.e}\onedot}

\def\wrt{w.r.t\onedot}

\def\iid{i.i.d\onedot}

\algblock{ParFor}{EndParFor}
\algnewcommand\algorithmicparfor{\textbf{parfor}}
\algnewcommand\algorithmicpardo{\textbf{do}}
\algnewcommand\algorithmicendparfor{\textbf{end\ parfor}}
\algrenewtext{ParFor}[1]{\algorithmicparfor\ #1\ \algorithmicpardo}
\algrenewtext{EndParFor}{\algorithmicendparfor}

\title{Improved Techniques for Training Score-Based Generative Models}

\author{%
 Yang Song \\
  Computer Science Department\\
  Stanford University \\
  \texttt{yangsong@cs.stanford.edu} \\
  \And
  Stefano Ermon\\
  Computer Science Department\\
  Stanford University\\
  \texttt{ermon@cs.stanford.edu}
}
\begin{document}
\maketitle
\begin{abstract}
Score-based generative models can produce high quality image samples comparable to GANs, without requiring adversarial optimization. However, existing training procedures are limited to images of low resolution (typically below $32\times 32$), and can be unstable under some settings. We provide a new theoretical analysis of learning and sampling from score-based models in high dimensional spaces, explaining existing failure modes and motivating new solutions that generalize across datasets. 
To enhance stability, we also propose to maintain an exponential moving average of model weights. With these improvements, we can scale score-based generative models to various image datasets, with diverse resolutions ranging from $64 \times 64$ to $25 6 \times 256$. Our score-based models can generate high-fidelity samples that rival best-in-class GANs on various image datasets, including CelebA, FFHQ, and several LSUN categories.

\end{abstract}
\section{Introduction}

Score-based generative models~\cite{song2019generative} represent probability distributions through score---a vector field pointing in the direction where the likelihood of data increases most rapidly. 
Remarkably, these score functions can be learned from data without requiring adversarial optimization, and can produce realistic image samples that rival GANs on simple datasets such as CIFAR-10~\cite{krizhevsky2009learning}.

Despite this success, existing score-based generative models only work on low resolution images ($32\times 32$) due to several limiting factors. 
First, the score function is learned via denoising score matching~\cite{hyvarinen2005estimation,vincent2011connection,raphan2011least}. Intuitively, this means a  neural network (named the \emph{score network}) is trained to denoise images blurred with Gaussian noise. A key insight from \cite{song2019generative} is to perturb the data using \emph{multiple} noise scales so that the score network captures both coarse and fine-grained image features. However, it is an open question how these noise scales should be chosen. The recommended settings in \cite{song2019generative} work well for $32\times 32$ images, but perform poorly when the resolution gets higher. Second, samples are generated by running Langevin dynamics~\cite{roberts1996exponential,welling2011bayesian}. This method starts from white noise and progressively denoises it into an image using the score network. This procedure, however, might fail or take an extremely long time to converge when used in high-dimensions and with a necessarily imperfect (learned) score network.

\begin{figure}
    \centering
    \includegraphics[width=\textwidth]{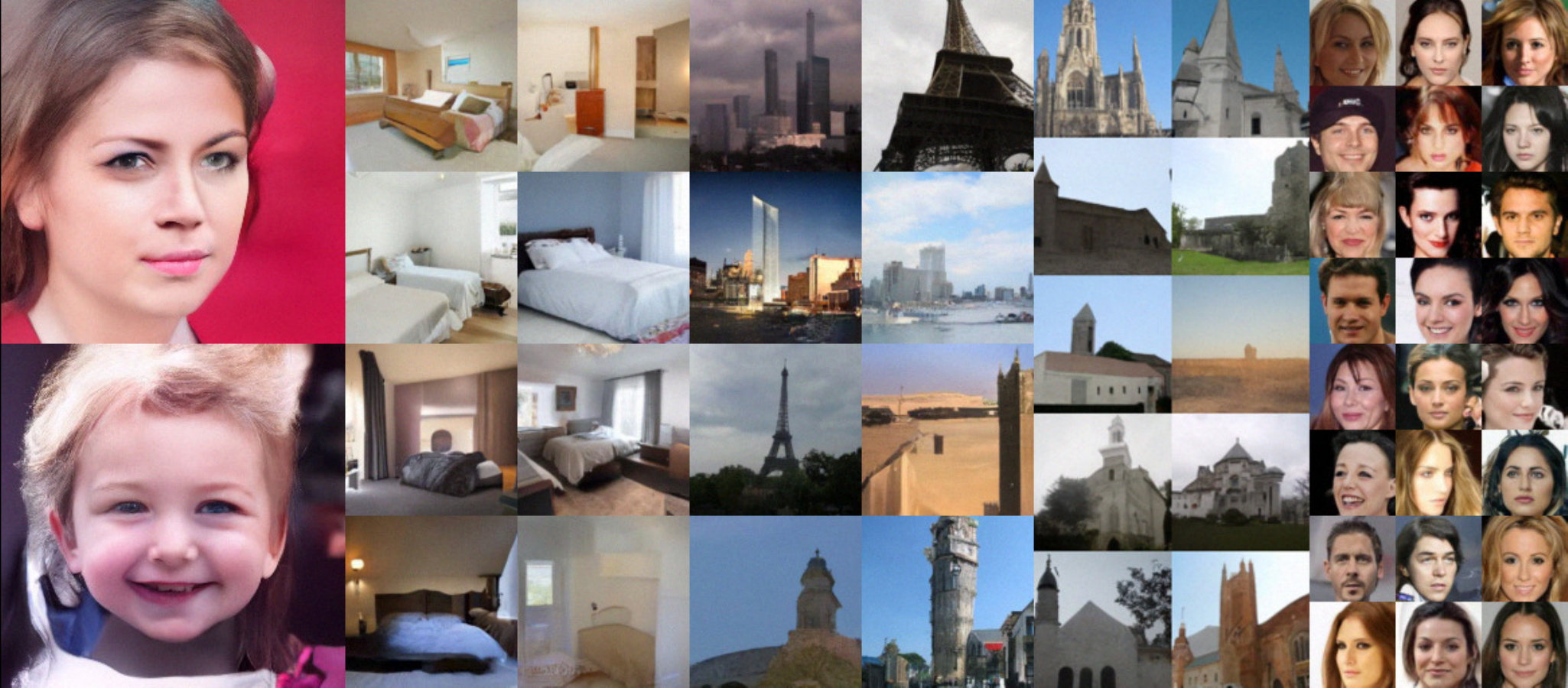}
    \caption{Generated samples on datasets of decreasing resolutions. From left to right: FFHQ $256\times 256$, LSUN bedroom $128\times 128$, LSUN tower $128\times 128$, LSUN church\_outdoor $96\times 96$, and CelebA $64\times 64$.}
    \label{fig:showcase}
\end{figure}

We propose a set of techniques to scale score-based generative models to high resolution images. Based on a new theoretical analysis on a simplified mixture model, we provide a method to analytically compute an effective set of Gaussian noise scales from training data. Additionally, we propose an efficient architecture to amortize the score estimation task across a large (possibly infinite) number of noise scales with a single neural network. Based on a simplified analysis of the convergence properties of the underlying Langevin dynamics sampling procedure, we also derive a technique to approximately optimize its performance as a function of the noise scales. 
Combining these techniques with an exponential moving average (EMA) of model parameters, we are able to significantly improve the sample quality, and successfully scale to images of resolutions ranging from $64\times 64$ to $256\times 256$, which was previously impossible for score-based generative models. As illustrated in \cref{fig:showcase}, the samples are sharp and diverse.

\section{Background}\label{sec:background}
\subsection{Langevin dynamics}
For any continuously differentiable probability density $p(\bfx)$, we call $\nabla_\bfx \log p(\bfx)$ its \emph{score function}. In many situations the score function is easier to model and estimate than the original probability density function~\cite{hyvarinen2005estimation, song2019sliced}. For example, for an unnormalized density it does not depend on the partition function. Once the score function is known, we can employ Langevin dynamics to sample from the corresponding distribution. Given a step size $\alpha > 0$, a total number of iterations $T$, and an initial sample $\bfx_0$ from any prior distribution $\pi(\bfx)$, Langevin dynamics iteratively evaluate the following
\begin{align}
    \bfx_{t} \gets \bfx_{t-1} + \alpha ~\nabla_\bfx \log p(\bfx_{t-1}) + \sqrt{2\alpha} ~\bfz_{t}, \quad 1 \le t \le T \label{eqn:langevin}
\end{align}
where $\bfz_{t} \sim \mcal{N}(\bfzero, \bfI)$. When $\alpha$ is sufficiently small and $T$ is sufficiently large, the distribution of $\bfx_T$ will be close to $p(\bfx)$ under some regularity conditions~\cite{roberts1996exponential,welling2011bayesian}. Suppose we have a neural network $\bfs_\bftheta(\bfx)$ (called the \emph{score network}) parameterized by $\bftheta$, and it has been trained such that $\bfs_\bftheta(\bfx) \approx \nabla_\bfx \log p(\bfx)$. We can approximately generate samples from $p(\bfx)$ using Langevin dynamics by replacing $\nabla_\bfx \log p(\bfx_{t-1})$ with $\bfs_{\bftheta}(\bfx_{t-1})$ in \cref{eqn:langevin}.
Note that \cref{eqn:langevin} can be interpreted as noisy gradient ascent on the log-density $\log p(\bfx)$.

\subsection{Score-based generative modeling}\label{sec:ald}
We can estimate the score function from data and generate new samples with Langevin dynamics. This idea was named \emph{score-based generative modeling} by ref.~\cite{song2019generative}. Because the estimated score function is inaccurate in regions without training data, Langevin dynamics may not converge correctly when a sampling trajectory encounters those regions (see more detailed analysis in ref.~\cite{song2019generative}). As a remedy, ref.~\cite{song2019generative} proposes to perturb the data with Gaussian noise of different intensities and jointly estimate the score functions of all noise-perturbed data distributions. During inference, they combine the information from all noise scales by sampling from each noise-perturbed distribution sequentially with Langevin dynamics.

More specifically, suppose we have an underlying data distribution $p_\text{data}(\bfx)$ and consider a sequence of noise scales $\{\sigma_i\}_{i=1}^L$ that satisfies $\sigma_1 > \sigma_2 > \cdots > \sigma_L$. Let $p_{\sigma}(\tilde{\bfx} \mid \bfx) = \mcal{N}(\tilde{\bfx} \mid \bfx, \sigma^2 \bfI)$, and denote the corresponding perturbed data distribution as $p_\sigma(\tilde{\bfx}) \triangleq \int p_\sigma(\tilde{\bfx} \mid \bfx) p_\text{data}(\bfx) \mathrm{d} \bfx$. Ref.~\cite{song2019generative} proposes to estimate the score function of each $p_{\sigma_i}(\bfx)$ by training a joint neural network $\bfs_\bftheta(\bfx, \sigma)$ (called the \emph{noise conditional score network}) with the following loss:
\begin{align}
  \frac{1}{2 L}   \sum_{i=1}^L \mbb{E}_{p_\text{data}(\bfx)} \mbb{E}_{p_{\sigma_i}(\tilde{\bfx}\mid\bfx)}\bigg[ \norm{\sigma_i \bfs_\bftheta(\tilde{\bfx}, \sigma_i) + \frac{\tilde{\bfx} - \bfx}{\sigma_i}}_2^2 \bigg] \label{eqn:ncsn_train},
\end{align}%
where all expectations can be efficiently estimated using empirical averages. When trained to the optimum (denoted as $s_{\bftheta^*}(\bfx, \sigma)$), the noise conditional score network (NCSN) satisfies $\forall i: s_{\bftheta^*}(\bfx, \sigma_i) = \nabla_\bfx \log p_{\sigma_i}(\bfx)$ almost everywhere~\cite{song2019generative}, assuming enough data and model capacity.%

\begin{wrapfigure}[15]{r}{0.5\textwidth}
\vspace{-2.2em}
\begin{minipage}{0.5\textwidth}
\begin{algorithm}[H]
	\caption{Annealed Langevin dynamics~\cite{song2019generative}}
	\label{alg:anneal}
	\begin{algorithmic}[1]
	    \Require{$\{\sigma_i\}_{i=1}^L, \epsilon, T$.}
	    \State{Initialize $\bfx_0$}
	    \For{$i \gets 1$ to $L$}
	        \State{$\alpha_i \gets \epsilon \cdot \sigma_i^2/\sigma_L^2$} \Comment{$\alpha_i$ is the step size.}
            \For{$t \gets 1$ to $T$}
                \State{Draw $\bfz_t \sim \mcal{N}(0, I)$}
                \State{\resizebox{0.75\textwidth}{!}{$\bfx_t \gets \bfx_{t-1} + \alpha_i~ \bfs_\bftheta(\bfx_{t-1}, \sigma_i) + \sqrt{2 \alpha_i}~ \bfz_{t}$}}
            \EndFor
            \State{$\bfx_0 \gets \bfx_T$}
        \EndFor
        \If{denoise $\bfx_T$}
        \State{\textbf{return} $\bfx_T + \sigma_T^2 \bfs_\bftheta(\bfx_T, \sigma_T)$}
        \Else
        \State{\textbf{return} $\bfx_T$}
        \EndIf
	\end{algorithmic}
\end{algorithm}
\end{minipage}
\end{wrapfigure}
After training an NCSN, ref.~\cite{song2019generative} generates samples by \emph{annealed Langevin dynamics}, a method that combines information from all noise scales. We provide its pseudo-code in \cref{alg:anneal}. The approach amounts to sampling from $p_{\sigma_1}(\bfx), p_{\sigma_2}(\bfx), \cdots, p_{\sigma_L}(\bfx)$ sequentially with Langevin dynamics with a special step size schedule $\alpha_i = \epsilon~ \sigma_i^2 / \sigma_L^2$ for the $i$-th noise scale. Samples from each noise scale are used to initialize Langevin dynamics for the next noise scale until reaching the smallest one, where it provides final samples for the NCSN.

Following the first public release of this work, ref.~\cite{jolicoeurpiche2020adversarial} noticed that adding an extra denoising step after the original annealed Langevin dynamics in \cite{song2019generative}, similar to \cite{saremi2019neural,li2019learning,kadkhodaie2020solving}, often significantly improves FID scores~\cite{heusel2017gans} without affecting the visual appearance of samples. Instead of directly returning $\bfx_T$, this denoising step returns $\bfx_T + \sigma_T^2 \bfs_\bftheta(\bfx_T, \sigma_T)$ (see \cref{alg:anneal}), which essentially removes the unwanted noise $\mcal{N}(\bfzero, \sigma_T^2 \bfI)$ from $\bfx_T$ using Tweedie's formula~\cite{efron2011tweedie}. Therefore, we have updated results in the main paper by incorporating this denoising trick, but kept some original results without this denoising step in the appendix for reference.

There are many design choices that are critical to the successful training and inference of NCSNs, including (i) the set of noise scales $\{\sigma_i\}_{i=1}^L$, (ii) the way that $\bfs_\bftheta(\bfx, \sigma)$ incorporates information of $\sigma$, (iii) the step size parameter $\epsilon$ and (iv) the number of sampling steps per noise scale $T$ in \cref{alg:anneal}. Below we provide theoretically motivated ways to configure them without manual tuning, which significantly improve the performance of NCSNs on high resolution images.

\section{Choosing noise scales}
Noise scales are critical for the success of NCSNs. As shown in \cite{song2019generative}, score networks trained with a single noise can never produce convincing samples for large images. Intuitively, high noise facilitates the estimation of score functions, but also leads to corrupted samples; while lower noise gives clean samples but makes score functions harder to estimate. One should therefore leverage different noise scales together to get the best of both worlds.

When the range of pixel values is $[0, 1]$, the original work on NCSN~\cite{song2019generative} recommends choosing $\{\sigma_i\}_{i=1}^L$ as a geometric sequence where $L = 10$, $\sigma_1 = 1$, and $\sigma_{L} = 0.01$. It is reasonable that the smallest noise scale $\sigma_L = 0.01 \ll 1$, because we sample from perturbed distributions with descending noise scales and we want to add low noise at the end.
However, some important questions remain unanswered, which turn out to be critical to the success of NCSNs on high resolution images: (i) Is $\sigma_1 = 1$ appropriate? If not, how should we adjust $\sigma_1$ for different datasets? (ii) Is geometric progression a good choice? (iii) Is $L=10$ good across different datasets? If not, how many noise scales are ideal? 

Below we provide answers to the above questions, motivated by theoretical analyses on simple mathematical models. Our insights are effective for configuring score-based generative modeling in practice, as corroborated by experimental results in \cref{sec:experiment}.

\subsection{Initial noise scale}
The algorithm of annealed Langevin dynamics (\cref{alg:anneal}) is an iterative refining procedure that starts from generating coarse samples with rich variation under large noise, before converging to fine samples with less variation under small noise. The initial noise scale $\sigma_1$ largely controls the diversity of the final samples. 
In order to promote sample diversity, we might want to choose $\sigma_1$ to be as large as possible. However, an excessively large $\sigma_1$ will require more noise scales (to be discussed in \cref{sec:noise_levels}) and make annealed Langevin dynamics more expensive. Below we present an analysis to guide the choice of $\sigma_1$ and provide a technique to strike the right balance.

Real-world data distributions are complex and hard to analyze, so we approximate them with empirical distributions. Suppose we have a dataset $\{\bfx^{(1)}, \bfx^{(2)}, \cdots, \bfx^{(N)}\}$ which is \iid sampled from $p_\text{data}(\bfx)$. Assuming $N$ is sufficiently large, we have $p_\text{data}(\bfx) \approx \hat{p}_{\text{data}}(\bfx) \triangleq \frac{1}{N}\sum_{i=1}^N \delta(\bfx = \bfx^{(i)})$, where $\delta(\cdot)$ denotes a point mass distribution. When perturbed with $\mcal{N}(\bfzero, \sigma_1^2 \bfI)$, the empirical distribution becomes $\hat{p}_{\sigma_1}(\bfx) \triangleq \frac{1}{N} \sum_{i=1}^N p^{(i)}(\bfx)$, where $p^{(i)}(\bfx) \triangleq \mcal{N}(\bfx \mid \bfx^{(i)}, \sigma_1^2 \bfI)$. For generating diverse samples regardless of initialization, we naturally expect that Langevin dynamics can explore any component $p^{(i)}(\bfx)$ when initialized from any other component $p^{(j)}(\bfx)$, where $i \neq j$. The performance of Langevin dynamics is governed by the score function $\nabla_\bfx \log \hat{p}_{\sigma_1}(\bfx)$ (see \cref{eqn:langevin}).

\begin{proposition}\label{prop:init_noise}
Let $\hat{p}_{\sigma_1}(\bfx) \triangleq \frac{1}{N} \sum_{i=1}^N p^{(i)}(\bfx)$, where $p^{(i)}(\bfx) \triangleq \mcal{N}(\bfx \mid \bfx^{(i)}, \sigma_1^2 \bfI)$. With $r^{(i)}(\bfx) \triangleq \frac{p^{(i)}(\bfx)}{\sum_{k=1}^N  p^{(k)}(\bfx) }$, the score function is $\nabla_\bfx \log \hat{p}_{\sigma_1}(\bfx) = \sum_{i=1}^N r^{(i)}(\bfx) \nabla_\bfx \log p^{(i)}(\bfx)$. Moreover,
\begin{align}
    \mbb{E}_{p^{(i)}(\bfx)}[r^{(j)}(\bfx)] \leq \frac{1}{2} \exp \bigg( -\frac{\norm{\bfx^{(i)} - \bfx^{(j)}}_2^2}{8\sigma_1^2} \bigg). \label{eqn:ratio}
\end{align}
\end{proposition}
In order for Langevin dynamics to transition from $p^{(i)}(\bfx)$ to $p^{(j)}(\bfx)$ easily for $i \neq j$, $\mbb{E}_{p^{(i)}(\bfx)}[r^{(j)}(\bfx)]$ has to be relatively large, because otherwise $\nabla_\bfx \log \hat{p}_{\sigma_1}(\bfx) = \sum_{k=1}^N r^{(k)}(\bfx) \nabla_\bfx \log p^{(k)}(\bfx)$ will ignore the component $p^{(j)}(\bfx)$ (on average) when initialized with $\bfx \sim p^{(i)}(\bfx)$ and in such case Langevin dynamics will act as if $p^{(j)}(\bfx)$ did not exist. The bound of \cref{eqn:ratio} indicates that $\mbb{E}_{p^{(i)}(\bfx)}[r^{(j)}(\bfx)]$ can decay exponentially fast if $\sigma_1$ is small compared to $\norm{\bfx^{(i)} - \bfx^{(j)}}_2$. As a result, it is necessary for $\sigma_1$ to be numerically comparable to the maximum pairwise distances of data to facilitate transitioning of Langevin dynamics and hence improving sample diversity. In particular, we suggest: 
\begin{recipe}[Initial noise scale]\label{rec:init_noise}
    Choose $\sigma_1$ to be as large as the maximum Euclidean distance between all pairs of training data points.%
\end{recipe}

\begin{wrapfigure}[12]{r}{0.5\textwidth}
    \vspace{-0.5em}
    \centering
    \begin{subfigure}[b]{0.15\textwidth}
        \includegraphics[width=\textwidth]{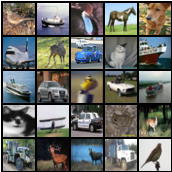}
        \caption{Data}\label{fig:init_noise_data}
    \end{subfigure}
    \begin{subfigure}[b]{0.15\textwidth}
        \includegraphics[width=\textwidth]{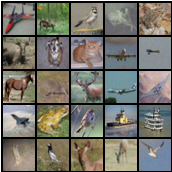}
        \caption{$\sigma_1 = 1$}\label{fig:init_noise_1}
    \end{subfigure}
    \begin{subfigure}[b]{0.15\textwidth}
        \includegraphics[width=\textwidth]{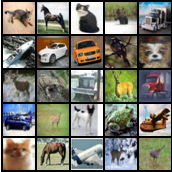}
        \caption{$\sigma_1 = 50$}\label{fig:init_noise_50}
    \end{subfigure}
    \caption{Running annealed Langevin dynamics to sample from a mixture of Gaussian centered at images in the CIFAR-10 test set.}
    \label{fig:init_noise}
\end{wrapfigure}
Taking CIFAR-10 as an example, the median pairwise distance between all training images is around 18, so $\sigma_1 = 1$ as in~\cite{song2019generative} implies $\mbb{E}[r(\bfx)] < 10^{-17}$ and is unlikely to produce diverse samples as per our analysis. To test whether choosing $\sigma_1$ according to \cref{rec:init_noise} (\ie, $\sigma_1 = 50$) gives significantly more diverse samples than using $\sigma_1 = 1$, we run annealed Langevin dynamics to sample from a mixture of Gaussian with 10000 components, where each component is centered at one CIFAR-10 test image. All initial samples are drawn from a uniform distribution over $[0,1]^{32\times32\times 3}$. This setting allows us to avoid confounders introduced by NCSN training because we use ground truth score functions. As shown in \cref{fig:init_noise}, samples in \cref{fig:init_noise_50} (using \cref{rec:init_noise}) exhibit comparable diversity to ground-truth images (\cref{fig:init_noise_data}), and have better variety than \cref{fig:init_noise_1} ($\sigma_1 = 1$). Quantitatively, the average pairwise distance of samples in \cref{fig:init_noise_50} is 18.65, comparable to data (17.78) but much higher than that of \cref{fig:init_noise_1} (10.12).

\subsection{Other noise scales}\label{sec:noise_levels}

After setting $\sigma_L$ and $\sigma_1$, we need to choose the number of noise scales $L$ and specify the other elements of $\{\sigma_i\}_{i=1}^L$. As analyzed in \cite{song2019generative}, it is crucial for the success of score-based generative models to ensure that $p_{\sigma_i}(\bfx)$ generates a sufficient number of training data in high density regions of $p_{\sigma_{i-1}}(\bfx)$ for all $1 < i \leq L$. The intuition is we need reliable gradient signals for $p_{\sigma_i}(\bfx)$ when initializing Langevin dynamics with samples from $p_{\sigma_{i-1}}(\bfx)$. 

However, an extensive grid search on $\{\sigma_i\}_{i=1}^L$ can be very expensive. To give some theoretical guidance on finding good noise scales, we consider a simple case where the dataset contains only one data point, or equivalently, $\forall 1\leq i \leq L: p_{\sigma_i}(\bfx) = \mcal{N}(\bfx\mid \bfzero, \sigma_i^2 \bfI)$. Our first step is to understand the distributions of $p_{\sigma_i}(\bfx)$ better, especially when $\bfx$ has high dimensionality. We can decompose $p_{\sigma_i}(\bfx)$ in hyperspherical coordinates to $p(\bfphi)p_{\sigma_i}(r)$, where $r$ and $\bfphi$ denote the radial and angular coordinates of $\bfx$ respectively. Because $p_{\sigma_i}(\bfx)$ is an isotropic Gaussian, the angular component $p(\bfphi)$ is uniform and shared across all noise scales. As for $p_{\sigma_i}(r)$, we have the following
\begin{proposition}\label{prop:noise_levels}
Let $\bfx \in \mbb{R}^D \sim \mcal{N}(\bfzero, \sigma^2 \bfI)$, and $r = \norm{\bfx}_2$. We have
\begin{align*}
    p(r) = \frac{1}{2^{D/2 - 1}\Gamma(D/2)} \frac{r^{D-1}}{\sigma^{D}} \exp \bigg(-\frac{r^2}{2\sigma^2} \bigg)\quad \text{and} \quad
    r - \sqrt{D}\sigma \stackrel{d}{\to} \mcal{N}(0, \sigma^2 / 2) ~~\text{when $D \to \infty$}.
\end{align*}
\end{proposition}
In practice, dimensions of image data can range from several thousand to millions, and are typically large enough to warrant $p(r) \approx \mcal{N}(r| \sqrt{D} \sigma, \sigma^2 / 2)$ with negligible error. We therefore take $p_{\sigma_i}(r) = \mcal{N}(r | m_i, s_i^2)$ to simplify our analysis, where $m_i \triangleq \sqrt{D}\sigma$, and $s_i^2 \triangleq \sigma^2 / 2$.

Recall that our goal is to make sure samples from $p_{\sigma_i}(\bfx)$ will cover high density regions of $p_{\sigma_{i-1}}(\bfx)$. Because $p(\bfphi)$ is shared across all noise scales, $p_{\sigma_i}(\bfx)$ already covers the angular component of $p_{\sigma_{i-1}}(\bfx)$. Therefore, we need the radial components of $p_{\sigma_i}(\bfx)$ and $p_{\sigma_{i-1}}(\bfx)$ to have large overlap. %
Since $p_{\sigma_{i-1}}(r)$ has high density in $\mcal{I}_{i-1} \triangleq [m_{i-1} - 3s_{i-1}, m_{i-1} + 3s_{i-1}]$ (employing the ``three-sigma rule of thumb''~\cite{grafarend2006linear}), a natural choice is to fix $p_{\sigma_{i}}(r \in \mcal{I}_{i-1}) = \Phi(\sqrt{2D}(\gamma_i - 1) + 3\gamma_i) - \Phi(\sqrt{2D}(\gamma_i - 1) - 3\gamma_i) = C$ with some moderately large constant $C > 0$ for all $1 < i \leq L$, where $\gamma_i \triangleq \sigma_{i-1}/\sigma_i$ and $\Phi(\cdot)$ is the CDF of standard Gaussian. This choice immediately implies that $\gamma_2 = \gamma_3 =\cdots \gamma_L$ and thus $\{\sigma_i\}_{i=1}^L$ is a geometric progression.

Ideally, we should choose as many noise scales as possible to make $C\approx 1$. However, having too many noise scales will make sampling very costly, as we need to run Langevin dynamics for each noise scale in sequence. On the other hand, $L=10$ (for $32\times 32$ images) as in the original setting of \cite{song2019generative} is arguably too small, for which $C=0$ up to numerical precision. To strike a balance, we recommend $C \approx 0.5$ which performs well in our experiments. In summary,
\begin{recipe}[Other noise scales]\label{rec:noise_levels}
    Choose $\{\sigma_i\}_{i=1}^L$ as a geometric progression with common ratio $\gamma$, such that $\Phi(\sqrt{2D}(\gamma - 1) + 3\gamma) - \Phi(\sqrt{2D}(\gamma - 1) - 3\gamma) \approx 0.5$.
\end{recipe}

\subsection{Incorporating the noise information}\label{sec:cond}
\begin{wrapfigure}[11]{r}{0.35\textwidth}
    \centering
    \vspace{-1.8em}
    \hspace{-1em}\includegraphics[width=0.35\textwidth]{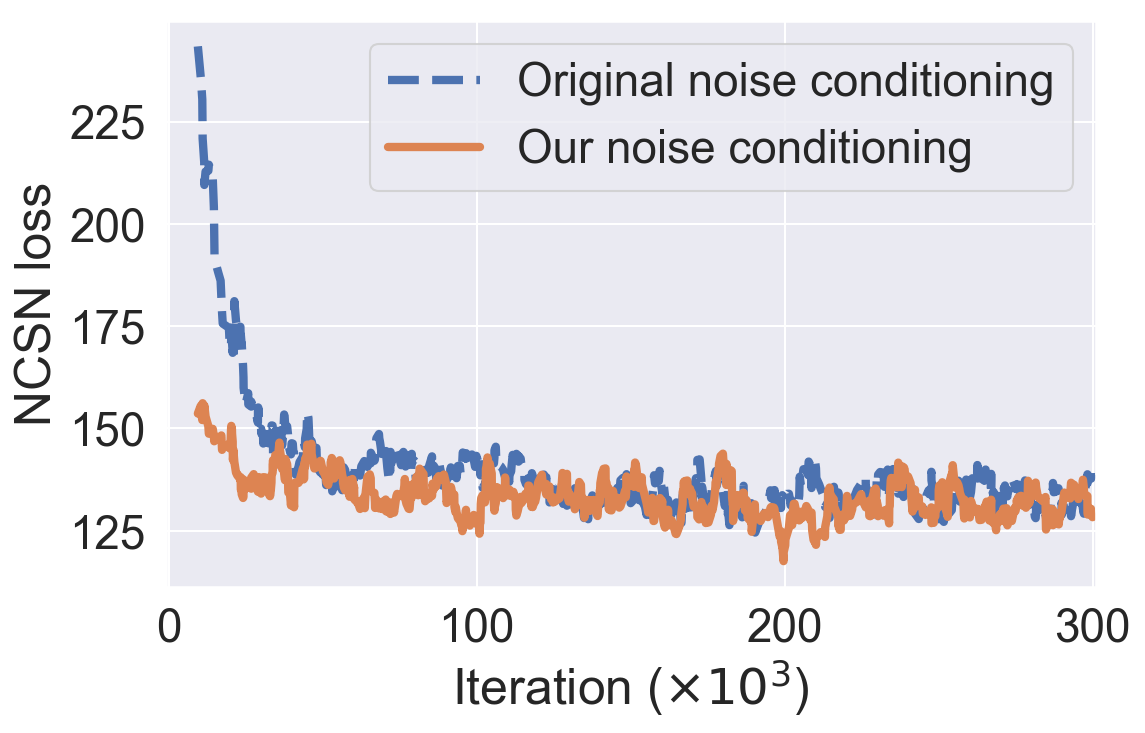}
    \caption{Training loss curves of two noise conditioning methods. }
    \label{fig:loss_compare}
\end{wrapfigure}
For high resolution images, we need a large $\sigma_1$ and a huge number of noise scales as per \cref{rec:init_noise} and \ref{rec:noise_levels}. Recall that the NCSN is a single amortized network that takes a noise scale and gives the corresponding score.
In~\cite{song2019generative}, authors use a separate set of scale and bias parameters in normalization layers to incorporate the information from each noise scale. However, its memory consumption grows linearly \wrt $L$, and it is not applicable when the NCSN has no normalization layers.

We propose an efficient alternative that is easier to implement and more widely applicable. For $p_\sigma(\bfx) = \mcal{N}(\bfx\mid \bfzero, \sigma^2 \bfI)$ analyzed in \cref{sec:noise_levels}, we observe that $\mbb{E}[\norm{\nabla_\bfx \log p_\sigma(\bfx)}_2] \approx \nicefrac{\sqrt{D}}{\sigma}$. Moreover, as empirically noted in \cite{song2019generative}, $\norm{\bfs_\bftheta(\bfx, \sigma)}_2 \propto \nicefrac{1}{\sigma}$ for a trained NCSN on real data. Because the norm of score functions scales inverse proportionally to $\sigma$, we can incorporate the noise information by rescaling the output of an unconditional score network $\bfs_\bftheta(\bfx)$ with $1/\sigma$. This motivates our following recommendation 
\begin{recipe}[Noise conditioning]\label{rec:cond}
Parameterize the NCSN with $\bfs_\bftheta(\bfx, \sigma) = \bfs_\bftheta(\bfx) / \sigma$, where $\bfs_\bftheta(\bfx)$ is an unconditional score network.
\end{recipe}%
It is typically hard for deep networks to automatically learn this rescaling, because $\sigma_1$ and $\sigma_L$ can differ by several orders of magnitude. This simple choice is easier to implement, and can easily handle a large number of noise scales (even continuous ones). As shown in \cref{fig:loss_compare} (detailed settings in \cref{app:exp_details}), it achieves similar training losses compared to the original noise conditioning approach in \cite{song2019generative}, and generate samples of better quality (see \cref{app:ablation}).

\section{Configuring annealed Langevin dynamics}

In order to sample from an NCSN with annealed Langevin dynamics, we need to specify the number of sampling steps per noise scale $T$ and the step size parameter $\epsilon$ in \cref{alg:anneal}. Authors of \cite{song2019generative} recommends $\epsilon=2 \times 10^{-5}$ and $T = 100$. It remains unclear how we should change $\epsilon$ and $T$ for different sets of noise scales. 

To gain some theoretical insight, we revisit the setting in \cref{sec:noise_levels} where the dataset has one point (\ie, $p_{\sigma_i}(\bfx) = \mcal{N}(\bfx \mid \bfzero, \sigma_i^2 \bfI)$). Annealed Langevin dynamics connect two adjacent noise scales $\sigma_{i-1} > \sigma_i$ by initializing the Langevin dynamics for $p_{\sigma_i}(\bfx)$ with samples obtained from $p_{\sigma_{i-1}}(\bfx)$. When applying Langevin dynamics to $p_{\sigma_i}(\bfx)$, we have $\bfx_{t+1} \gets \bfx_t + \alpha \nabla_\bfx \log p_{\sigma_i}(\bfx_t) + \sqrt{2\alpha}\bfz_t$, where $\bfx_0 \sim p_{\sigma_{i-1}}(\bfx)$ and $\bfz_t \sim \mcal{N}(\bfzero, \bfI)$. The distribution of $\bfx_T$ can be computed in closed form:
\begin{proposition}\label{prop:langevin}
Let $\gamma = \frac{\sigma_{i-1}}{\sigma_i}$. For $\alpha = \epsilon\cdot  \frac{\sigma_i^2}{\sigma_L^2}$ (as in \cref{alg:anneal}), we have $\bfx_T \sim \mcal{N}(\bfzero, s^2_T \bfI)$, where
\begin{align}
    \frac{s^2_T}{\sigma_i^2} =  \bigg( 1 - \frac{\epsilon}{\sigma_L^2} \bigg)^{2T}\Bigg( \gamma^2 - \frac{2\epsilon}{\sigma_L^2 - \sigma_L^2 \left(1 - \frac{\epsilon}{\sigma_L^2} \right)^2} \Bigg) + \frac{2\epsilon}{\sigma_L^2 - \sigma_L^2 \left( 1 - \frac{\epsilon}{\sigma_L^2} \right)^2}.\label{eqn:step_size}
\end{align}
\end{proposition}
When $\{\sigma_i\}_{i=1}^L$ is a geometric progression as advocated by \cref{rec:noise_levels}, we immediately see that $\nicefrac{s^2_T}{\sigma_i^2}$ is identical across all $1 < i \leq T$ because of the shared $\gamma$. Furthermore, the value of $\nicefrac{s^2_T}{\sigma_i^2}$ has no explicit dependency on the dimensionality $D$. %

For better mixing of annealed Langevin dynamics, we hope $\nicefrac{s^2_T}{\sigma_i^2}$ 
approaches 1 across all noise scales, which can be achieved by finding $\epsilon$ and $T$ that minimize the difference between \cref{eqn:step_size} and 1. Unfortunately, this often results in an unnecessarily large $T$ that makes sampling very expensive for large $L$. As an alternative, we propose to first choose $T$ based on a reasonable computing budget (typically $T \times L$ is several thousand), and subsequently find $\epsilon$ by making \cref{eqn:step_size} as close to 1 as possible.
In summary:
\begin{recipe}[selecting $T$ and $\epsilon$]\label{rec:langevin}
    Choose $T$ as large as allowed by a computing budget and then select an $\epsilon$ that makes \cref{eqn:step_size} maximally close to 1.
\end{recipe}
We follow this guidance to generate all samples in this paper, except for those from the original NCSN where we adopt the same settings as in \cite{song2019generative}. When finding $\epsilon$ with \cref{rec:langevin} and \cref{eqn:step_size}, we recommend performing grid search over $\epsilon$, rather than using gradient-based optimization methods.

\section{Improving stability with moving average}\label{sec:stability}

\begin{figure}%
    \centering
    \begin{subfigure}[b]{0.43\textwidth}
        \includegraphics[width=\textwidth]{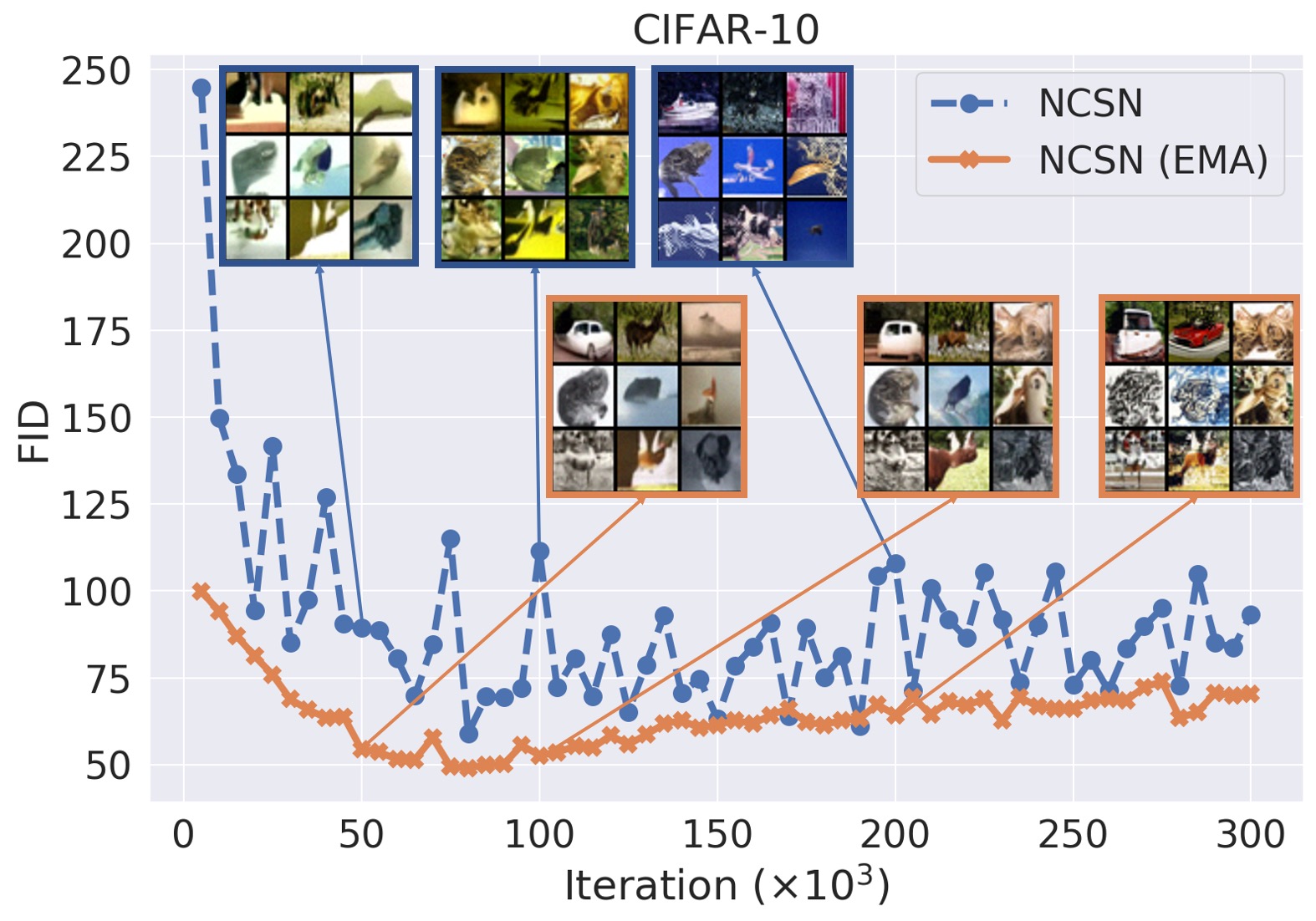}
    \end{subfigure}
    \begin{subfigure}[b]{0.43\textwidth}
        \includegraphics[width=\textwidth]{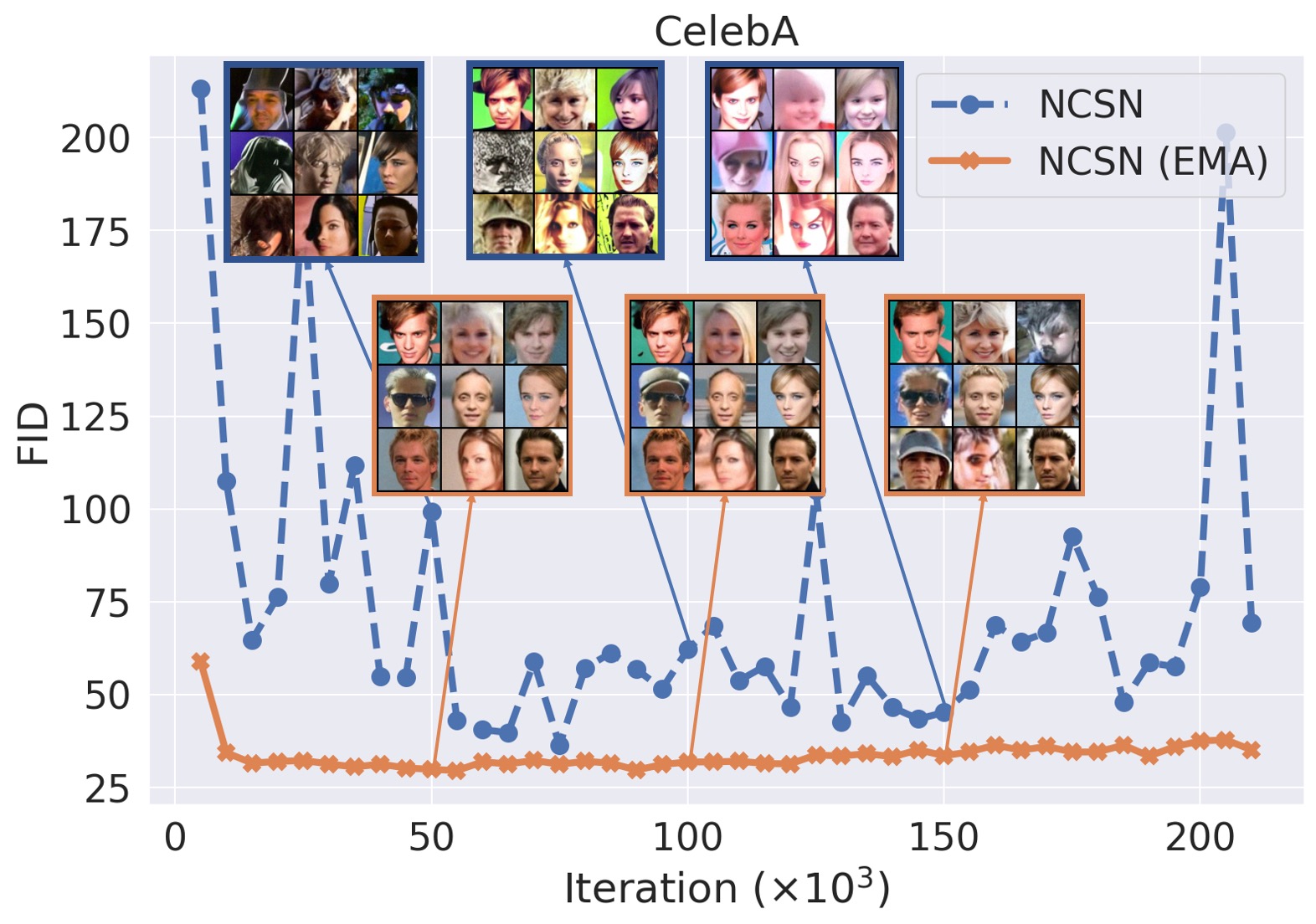}
    \end{subfigure}
    \caption{FIDs and color artifacts over the course of training (best viewed in color). The FIDs of NCSN have much higher volatility compared to NCSN with EMA. Samples from the vanilla NCSN often have obvious color shifts. All FIDs are computed with the denoising step.}
    \label{fig:stability}
\end{figure}

Unlike GANs, score-based generative models have one unified objective (\cref{eqn:ncsn_train}) and require no adversarial training.
However, even though the loss function of NCSNs typically decreases steadily over the course of training, we observe that the generated image samples sometimes exhibit unstable visual quality, especially for images of larger resolutions. We empirically demonstrate this fact by training NCSNs on CIFAR-10 $32\times 32$ and CelebA~\cite{liu2015faceattributes} $64\times 64$ following the settings of \cite{song2019generative}, which exemplifies typical behavior on other image datasets. We report FID scores~\cite{heusel2017gans} computed on 1000 samples every 5000 iterations. Results in \cref{fig:stability} are computed with the denoising step, but results without the denoising step are similar (see \cref{fig:stability_no_denosing} in \cref{app:no_denoise_results}). As shown in \cref{fig:stability,fig:stability_no_denosing}, the FID scores for the vanilla NCSN often fluctuate significantly during training. Additionally, samples from the vanilla NCSN sometimes exhibit characteristic artifacts: image samples from the same checkpoint have strong tendency to have a common color shift. Moreover, samples are shifted towards different colors throughout training. We provide more samples in \cref{app:color} to manifest this artifact.

This issue can be easily fixed by exponential moving average (EMA). %
Specifically, let $\bftheta_i$ denote the parameters of an NCSN after the $i$-th training iteration, and $\bftheta'$ be an independent copy of the parameters. We update $\bftheta'$ with $\bftheta' \gets m \bftheta' + (1 - m) \bftheta_i$ after each optimization step, where $m$ is the momentum parameter and typically $m=0.999$. When producing samples, we use $\bfs_{\bftheta'}(\bfx, \sigma)$ instead of $\bfs_{\bftheta_i}(\bfx, \sigma)$. As shown in \cref{fig:stability}, EMA can effectively stabilize FIDs, remove artifacts (more samples in \cref{app:color}) and give better FID scores in most cases. Empirically, we observe the effectiveness of EMA is universal across a large number of different image datasets. As a result, we recommend the following rule of thumb:
\begin{recipe}[EMA]\label{rec:ema}
Apply exponential moving average to parameters when sampling.
\end{recipe}

\section{Combining all techniques together}\label{sec:experiment}
Employing \cref{rec:init_noise}--\ref{rec:ema}, we build NCSNs that can readily work across a large number of different datasets, including high resolution images that were previously 
out of reach
with score-based generative modeling. 
Our modified model is named NCSNv2. For a complete description on experimental details and more results, please refer to \cref{app:exp_details} and \ref{app:exp_results}.

\begin{figure}
    \centering
    \begin{minipage}{0.53\textwidth}
        \centering
            \begin{subfigure}[b]{0.5\textwidth}
            \includegraphics[width=\textwidth]{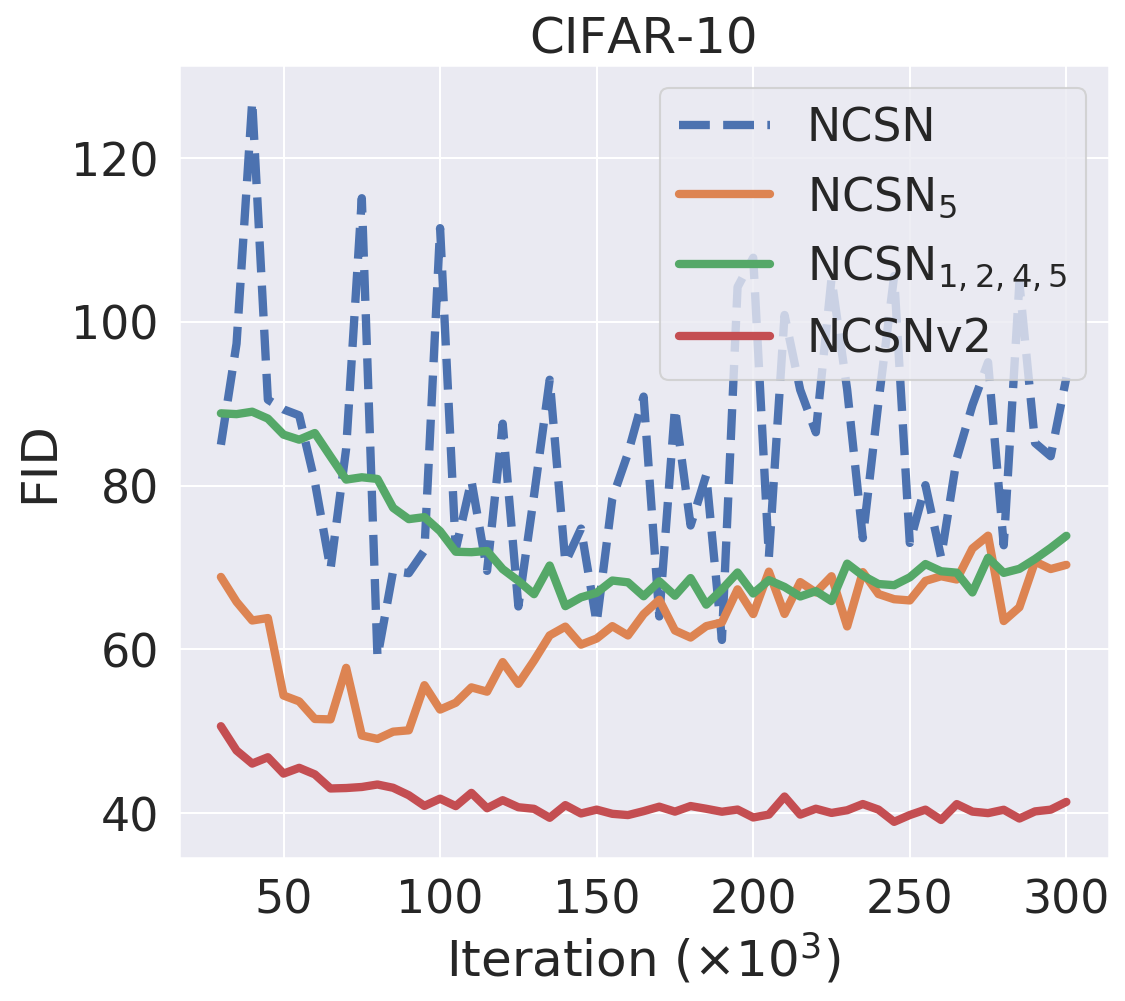}
            \caption{CIFAR-10 FIDs}\label{fig:fid_cifar10}
        \end{subfigure}%
        \begin{subfigure}[b]{0.5\textwidth}
            \includegraphics[width=\textwidth]{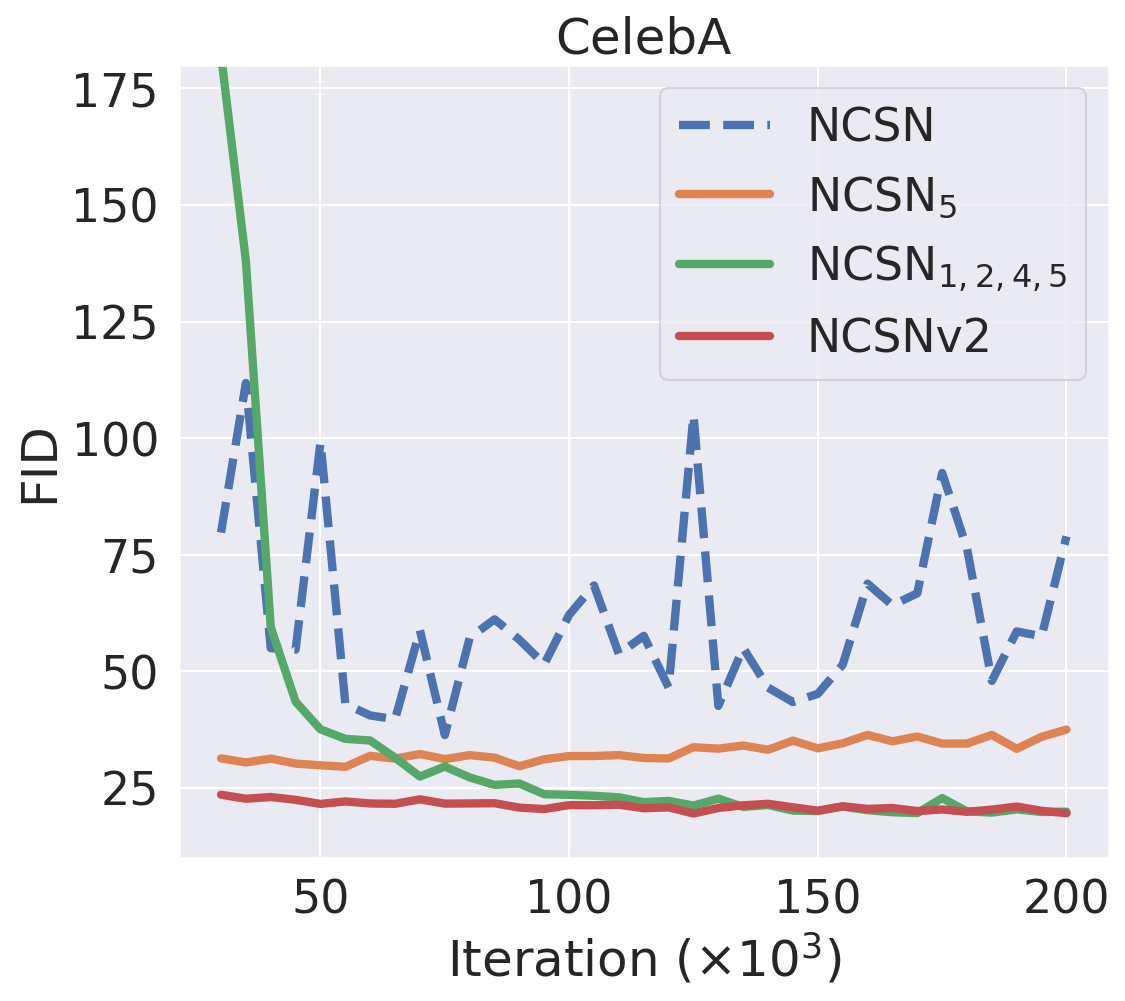}
            \caption{CelebA FIDs}\label{fig:fid_celeba}
        \end{subfigure}
        \caption{FIDs for different groups of techniques. Subscripts of ``NCSN'' are IDs of techniques in effect. ``NCSNv2'' uses all techniques. Results are computed with the denoising step.}
        \label{fig:ablation}
    \end{minipage}
    \begin{minipage}{0.45\textwidth}
    \captionof{table}{Inception and FID scores.} \label{tab:full_scores}
    \vspace{-0.5em}
       \centering
            \begin{adjustbox}{max width=0.9\linewidth}
            \begin{tabular}{lcc}
                \toprule
                Model & Inception $\uparrow$ & FID $\downarrow$\\
                \midrule
                \multicolumn{3}{l}{\textbf{CIFAR-10 Unconditional}} \\
                \midrule
                PixelCNN~\cite{van2016conditional} & $4.60$ & $65.93$\\
                IGEBM~\cite{du2019implicit} & $6.02$ & $40.58$ \\
                WGAN-GP~\cite{gulrajani2017improved} & $7.86 \pm .07$ & $36.4$\\
                SNGAN~\cite{miyato2018spectral} & $8.22\pm .05$ & $21.7$ \\
                \midrule
                NCSN~\cite{song2019generative} & $\mathbf{8.87 \pm .12}$ & $25.32$\\
                NCSN (w/ denoising) & $7.32 \pm .12$ & $29.8$\\
                NCSNv2 (w/o denoising) & $8.73 \pm .13$ & $31.75$\\
                NCSNv2 (w/ denoising) & $8.40 \pm .07$ & $\mathbf{10.87}$\\
                \midrule
                \multicolumn{3}{l}{\textbf{CelebA $\mathbf{64\times 64}$}}\\
                \midrule
                NCSN (w/o denoising) & - & $26.89$ \\
                NCSN (w/ denoising) & - & $25.30$\\
                NCSNv2 (w/o denoising) & - & $28.86$\\
                NCSNv2 (w/ denoising) & - & $\mathbf{10.23}$\\
                \bottomrule
            \end{tabular} 
            \end{adjustbox}
    \end{minipage}
\end{figure}

\textbf{Quantitative results:} We consider CIFAR-10 $32\times 32$ and CelebA $64\times 64$ where NCSN and NCSNv2 both produce reasonable samples. We report FIDs (lower is better) every 5000 iterations of training on 1000 samples and give results in \cref{fig:ablation} (with denoising) and \cref{fig:ablation_no_denoising} (without denoising, deferred to \cref{app:no_denoise_results}). As shown in \cref{fig:ablation,fig:ablation_no_denoising}, we observe that the FID scores of NCSNv2 (with all techniques applied) are on average better than those of NCSN, and have much smaller variance over the course of training. 
Following \cite{song2019generative}, we select checkpoints with the smallest FIDs (on 1000 samples) encountered during training, and compute full FID and Inception scores on more samples from them. As shown by results in \cref{tab:full_scores}, NCSNv2 (w/ denoising) is able to significantly improve the FID scores of NCSN on both CIFAR-10 and CelebA, while bearing a slight loss of Inception scores on CIFAR-10. However, we note that Inception and FID scores have known issues~\cite{barratt2018note,sajjadi2018assessing} and they should be interpreted with caution as they may not correlate with visual quality in the expected way. In particular, they can be sensitive to slight noise perturbations~\cite{razavi2019generating}, as shown by the difference of scores with and without denoising in \cref{tab:full_scores}. To verify that NCSNv2 indeed generates better images than NCSN, we provide additional uncurated samples in \cref{app:ablation} for visual comparison.

\textbf{Ablation studies:} We conduct ablation studies to isolate the contributions of different techniques. We partition all techniques into three groups: (i) \cref{rec:ema}, (ii) \cref{rec:init_noise},\ref{rec:noise_levels},\ref{rec:langevin}, and (iii) \cref{rec:cond}, where different groups can be applied simultaneously. \cref{rec:init_noise},\ref{rec:noise_levels} and \ref{rec:langevin} are grouped together because \cref{rec:init_noise} and \ref{rec:noise_levels} collectively determine the set of noise scales, and to sample from NCSNs trained with these noise scales we need \cref{rec:langevin} to configure annealed Langevin dynamics properly. We test the performance of successively removing groups (iii), (ii), (i) from NCSNv2, and report results in \cref{fig:ablation} for sampling with denoising and in \cref{fig:ablation_no_denoising} (\cref{app:no_denoise_results}) for sampling without denoising. All groups of techniques improve over the vanilla NCSN. Although the FID scores are not strictly increasing when removing (iii), (ii), and (i) progressively, we note that FIDs may not always correlate with sample quality well. In fact, we do observe decreasing sample quality by visual inspection (see \cref{app:ablation}), and combining all techniques gives the best samples.

\begin{figure}
    \centering
    \includegraphics[width=\textwidth]{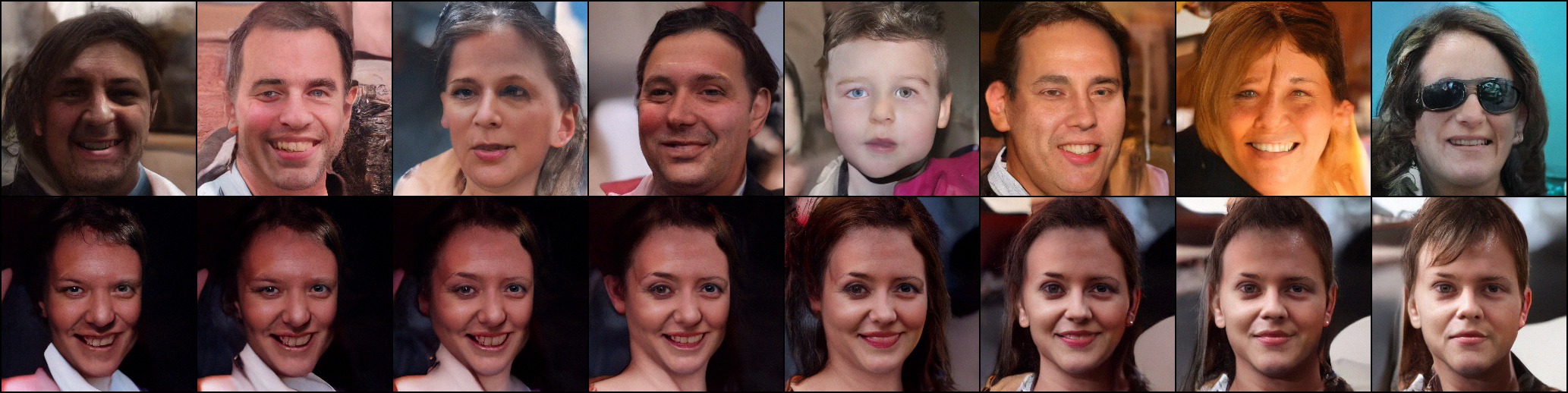}\\
    \includegraphics[width=\textwidth]{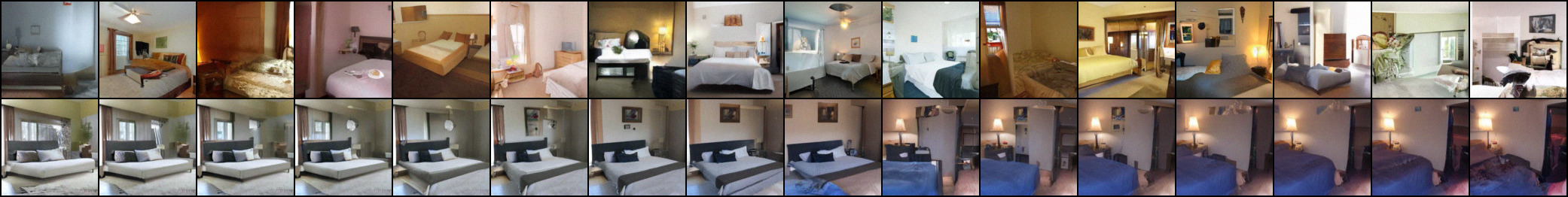}\\
    \includegraphics[width=\textwidth]{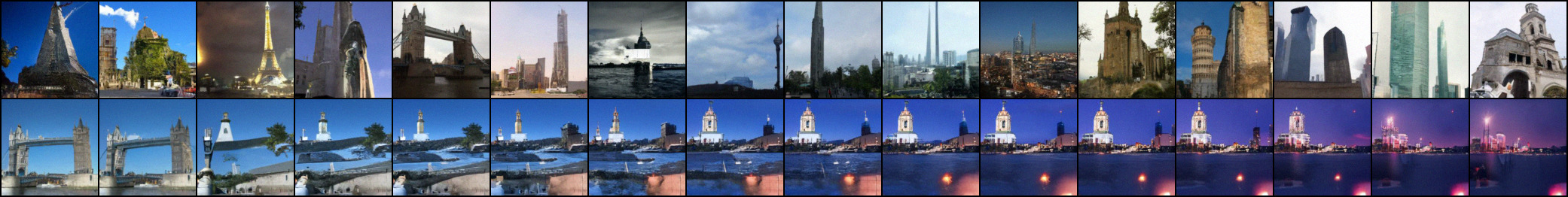}\\
    \includegraphics[width=\textwidth]{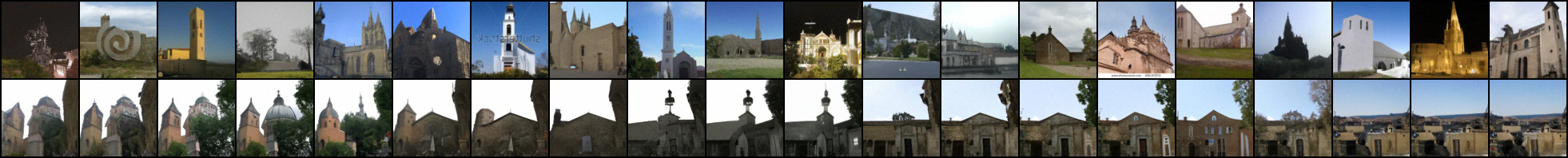}
    \caption{From top to bottom: FFHQ $256^2$, LSUN bedroom $128^2$, LSUN tower $128^2$, and LSUN church\_outdoor $96^2$. Within each group of images: the first row shows uncurated samples from NCSNv2, and the second shows the interpolation results between the leftmost and rightmost samples with NCSNv2. You may zoom in to view more details.}
    \label{fig:large_samples}
\end{figure}
\begin{figure}
    \centering
    \begin{subfigure}[b]{0.23\textwidth}
        \includegraphics[width=\textwidth]{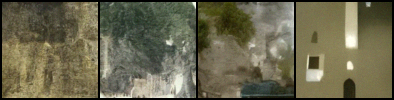}
        \caption{NCSN}
    \end{subfigure}
    \begin{subfigure}[b]{0.23\textwidth}
        \includegraphics[width=\textwidth]{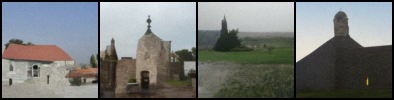}
        \caption{NCSNv2}
    \end{subfigure}
    \begin{subfigure}[b]{0.23\textwidth}
        \includegraphics[width=\textwidth]{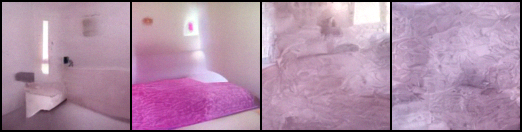}
        \caption{NCSN}
    \end{subfigure}
    \begin{subfigure}[b]{0.23\textwidth}
        \includegraphics[width=\textwidth]{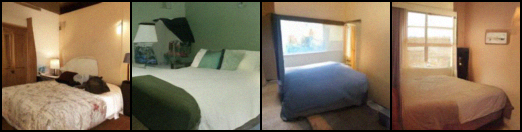}
        \caption{NCSNv2}
    \end{subfigure}
    \caption{NCSN vs. NCSNv2 samples on LSUN church\_outdoor (a)(b) and LSUN bedroom (c)(d).}
    \label{fig:sample_compare}
\end{figure}
\textbf{Towards higher resolution:} The original NCSN only succeeds at generating images of low resolution. In fact, \cite{song2019generative} only tested it on MNIST $28\times 28$ and CelebA/CIFAR-10 $32\times 32$. For slightly larger images such as CelebA $64\times 64$, NCSN can generate images of consistent global structure, yet with strong color artifacts that are easily noticeable (see \cref{fig:stability} and compare \cref{fig:celeba_ncsn_no_denoising} with \cref{fig:celeba_ncsnv2_no_denoising}). For images with resolutions beyond $96\times 96$, NCSN will completely fail to produce samples with correct structure or color (see \cref{fig:sample_compare}). All samples shown here are generated without the denoising step, but since $\sigma_L$ is very small, they are visually indistinguishable from ones with the denoising step.

By combining \cref{rec:init_noise}--\ref{rec:ema}, NCSNv2 can work on images of much higher resolution. Note that we directly calculated the noise scales for training NCSNs, and computed the step size for annealed Langevin dynamics sampling without manual hyper-parameter tuning. The network architectures are the same across datasets, except that for ones with higher resolution we use more layers and more filters to ensure the receptive field and model capacity are large enough (see details in \cref{app:arch}). In \cref{fig:large_samples} and \ref{fig:showcase}, we show NCSNv2 is capable of generating high-fidelity image samples with resolutions ranging from $96\times 96$ to $256 \times 256$. To show that this high sample quality is not a result of dataset memorization, we provide the loss curves for training/test, as well as nearest neighbors for samples in \cref{app:generalization}. In addition, NCSNv2 can produce smooth interpolations between two given samples as in \cref{fig:large_samples} (details in \cref{app:settings}), indicating the ability to learn generalizable image representations.

\section{Conclusion}
Motivated by both theoretical analyses and empirical observations, we propose a set of techniques to improve score-based generative models. Our techniques significantly improve the training and sampling processes,
lead to better sample quality, and enable high-fidelity image generation at high resolutions. Although our techniques work well without manual tuning, we believe that the performance can be improved even more by fine-tuning various hyper-parameters. Future directions include theoretical understandings on the sample quality of score-based generative models, as well as alternative noise distributions to Gaussian perturbations.

\section*{Broader Impact}
Our work represents another step towards more powerful generative models. While we focused on images, it is quite likely that similar techniques could be applicable to other data modalities such as speech or behavioral data (in the context of imitation learning). Like other generative models that have been previously proposed, such as GANs and WaveNets, score models have a multitude of applications. Among many other applications, they could be used to synthesize new data automatically, detect anomalies and adversarial examples, and also improve results in key tasks such as semi-supervised learning and reinforcement learning. In turn, these techniques can have both positive and negative impacts on society, depending on the application. In particular, the models we trained on image datasets can be used to synthesize new images that are hard to distinguish from real ones by humans. Synthetic images from generative models have already been used to deceive humans in malicious ways. There are also positive uses of these technologies, for example in the arts and as a tool to aid design in engineering. We also note that our models have been trained on datasets that have biases (\eg, CelebA is not gender-balanced), and the learned distribution is likely to have inherited them, in addition to others that are caused by the so-called inductive bias of models.

\section*{Acknowledgments and Disclosure of Funding}
The authors would like to thank Aditya Grover, Rui Shu and Shengjia Zhao for reviewing an early draft of this paper, as well as Gabby Wright and Sharon Zhou for resolving technical issues in computing HYPE$_\infty$ scores. This research was supported by NSF (\#1651565, \#1522054, \#1733686), ONR  (N00014-19-1-2145), AFOSR (FA9550-19-1-0024), and Amazon AWS.

\bibliography{ncsn2}
\bibliographystyle{unsrt}

\clearpage
\appendix
\section{Proofs}\label{app:proofs}
\begin{customprop}{\ref{prop:init_noise}}
Let $\hat{p}_{\sigma_1}(\bfx) \triangleq \frac{1}{N} \sum_{i=1}^N p^{(i)}(\bfx)$, where $p^{(i)}(\bfx) \triangleq \mcal{N}(\bfx \mid \bfx^{(i)}, \sigma_1^2 I)$. With $r^{(i)}(\bfx) \triangleq \frac{p^{(i)}(\bfx)}{\sum_{k=1}^N  p^{(k)}(\bfx) }$, the score function is $\nabla_\bfx \log \hat{p}_{\sigma_1}(\bfx) = \sum_{i=1}^N r^{(i)}(\bfx) \nabla_\bfx \log p^{(i)}(\bfx)$. Moreover,
\begin{align}
    \mbb{E}_{p^{(i)}(\bfx)}[r^{(j)}(\bfx)] \leq \frac{1}{2} \exp \bigg( -\frac{\norm{\bfx^{(i)} - \bfx^{(j)}}_2^2}{8\sigma_1^2} \bigg). \label{eqn:ratio2}
\end{align}
\end{customprop}
\begin{proof}
According to the definition of $p_{\sigma_1}(\bfx)$ and $r(\bfx)$, we have
\begin{align*}
    \nabla_\bfx \log \hat{p}_{\sigma_1}(\bfx) &= \nabla_\bfx \log \bigg( \frac{1}{N} \sum_{i=1}^N p^{(i)}(\bfx) \bigg)
    = \sum_{i=1}^N \frac{\nabla_\bfx p^{(i)}(\bfx)}{\sum_{j=1}^N p^{(j)}(\bfx)}\\
    &= \sum_{i=1}^N \frac{p^{(i)}(\bfx) \nabla_\bfx \log p^{(i)}(\bfx)}{\sum_{j=1}^N p^{(j)}(\bfx)} \\
    &= \sum_{i=1}^N r^{(i)}(\bfx) \nabla_\bfx \log p^{(i)}(\bfx).
\end{align*}
Next, assuming $\bfx \in \mbb{R}^D$, we have
\begin{align*}
    &\mbb{E}_{p^{(i)}(\bfx)}[r^{(j)}(\bfx)] = \int \frac{p^{(i)}(\bfx)p^{(j)}(\bfx)}{\sum_{k=1}^N p^{(k)}(\bfx)} \ud \bfx \leq \int \frac{p^{(i)}(\bfx)p^{(j)}(\bfx)}{p^{(i)}(\bfx) + p^{(j)}(\bfx)} \ud \bfx \\
    =& \frac{1}{2} \int \frac{2}{\frac{1}{p^{(i)}(\bfx)} + \frac{1}{p^{(j)}(\bfx)}} \ud \bfx \stackrel{(1)}{\leq} \frac{1}{2} \int \sqrt{p^{(i)}(\bfx) p^{(j)}(\bfx)} \ud \bfx\\
    =& \frac{1}{2} \frac{1}{(2\pi \sigma_1^2)^{D/2}} \int \exp \bigg(-\frac{1}{4\sigma_1^2}\bigg(\norm{\bfx - \bfx^{(i)}}_2^2 + \norm{\bfx - \bfx^{(j)}}_2^2\bigg) \bigg) \ud \bfx\\
    =& \frac{1}{2} \frac{1}{(2\pi \sigma_1^2)^{D/2}} \int \exp \bigg(-\frac{1}{4\sigma_1^2}\bigg(\norm{\bfx - \bfx^{(i)}}_2^2 + \norm{\bfx - \bfx^{(i)} + \bfx^{(i)} - \bfx^{(j)}}_2^2\bigg) \bigg) \ud \bfx\\
    =& \frac{1}{2} \frac{1}{(2\pi \sigma_1^2)^{D/2}} \int \exp \bigg\{-\frac{1}{2\sigma_1^2}\bigg(\norm{\bfx - \bfx^{(i)}}_2^2 + (\bfx - \bfx^{(i)})\tran (\bfx^{(i)} - \bfx^{(j)}) + \frac{\norm{\bfx^{(i)} - \bfx^{(j)}}_2^2}{2}\bigg) \bigg\} \ud \bfx\\
    =& \frac{1}{2} \frac{1}{(2\pi \sigma_1^2)^{D/2}} \int \exp \bigg\{-\frac{1}{2\sigma_1^2}\bigg(\norm{\bfx - \bfx^{(i)} + \frac{\bfx^{(i)} - \bfx^{(j)}}{2}}_2^2 + \frac{\norm{\bfx^{(i)} - \bfx^{(j)}}_2^2}{4}\bigg) \bigg\} \ud \bfx\\
    =& \frac{1}{2}\exp\bigg(-\frac{\norm{\bfx^{(i)} - \bfx^{(j)}}_2^2}{8\sigma_1^2}\bigg) \frac{1}{(2\pi \sigma_1^2)^{D/2}} \int \exp \bigg\{-\frac{1}{2\sigma_1^2}\bigg(\norm{\bfx - \bfx^{(i)} + \frac{\bfx^{(i)} - \bfx^{(j)}}{2}}_2^2\bigg) \bigg\} \ud \bfx\\
    =& \frac{1}{2}\exp\bigg(-\frac{\norm{\bfx^{(i)} - \bfx^{(j)}}_2^2}{8\sigma_1^2}\bigg),
\end{align*}
where $(1)$ is due to the geometric mean--harmonic mean inequality.
\end{proof}

\begin{customprop}{\ref{prop:noise_levels}}
Let $\bfx \in \mbb{R}^D \sim \mcal{N}(\bfzero, \sigma^2 I)$, and $r = \norm{\bfx}_2$. We have
\begin{align*}
    p(r) = \frac{1}{2^{D/2 - 1}\Gamma(D/2)} \frac{r^{D-1}}{\sigma^{D}} \exp \bigg(-\frac{r^2}{2\sigma^2} \bigg)\quad \text{and} \quad
    r - \sqrt{D}\sigma \stackrel{d}{\to} \mcal{N}(0, \sigma^2 / 2) ~~\text{when $D \to \infty$}.
\end{align*}
\end{customprop}
\begin{proof}
Since $\bfx \sim \mcal{N}(\bfzero, \sigma^2 I)$, we have $s \triangleq \norm{\bfx}_2^2 / \sigma^2 \sim \chi^2_D$, \ie,
\begin{align*}
    p_s(s) = \frac{1}{2^{D/2} \Gamma(D / 2)} s^{D/2 - 1} e^{-s / 2}.
\end{align*}
Because $r = \norm{\bfx}_2 = \sigma \sqrt{s}$, we can use the change of variables formula to get
\begin{align*}
    p(r) = \frac{2r}{\sigma^2} p_s(s) = \frac{2r}{\sigma^2} p_s\bigg(\frac{r^2}{\sigma^2}\bigg) = \frac{1}{2^{D/2 - 1}\Gamma(D/2)} \frac{r^{D-1}}{\sigma^{D}} \exp \bigg(-\frac{r^2}{2\sigma^2} \bigg),
\end{align*}
which proves our first result. Next, we notice that if $x \sim \mcal{N}(0, \sigma^2)$, we have $x^2/\sigma^2 \sim \chi^2_1$ and thus $\mbb{E}[x] = \sigma^2$, $\operatorname{Var}[x] = 2\sigma^4$.
As a result, if ${x_1, x_2, \cdots, x_D} \stackrel{\text{\iid}}{\sim} \mcal{N}(0, \sigma^2)$, the law of large numbers and the central limit theorem will imply that as $D \to \infty$, both of the following hold:
\begin{align*}
    \frac{x_1^2 + x_2^2 + \cdots + x_D^2}{D} &\stackrel{p}{\to} \sigma^2\\
    \sqrt{D}\bigg( \frac{x_1^2 + x_2^2 + \cdots + x_D^2}{D} - \sigma^2 \bigg) &\stackrel{d}{\to} \mcal{N}(0, 2\sigma^4).
\end{align*}
Equivalently,
\begin{align*}
    \sqrt{D}\bigg( \frac{r^2}{D} - \sigma^2 \bigg) \stackrel{d}{\to} \mcal{N}(0, 2\sigma^4).
\end{align*}
Applying the delta method, we obtain
\begin{align*}
    \sqrt{D} \bigg ( \frac{r}{\sqrt{D}} - \sigma \bigg) \stackrel{d}{\to} \mcal{N}(0, \sigma^2 / 2),
\end{align*}
and therefore $r - \sqrt{D}\sigma \stackrel{d}{\to} \mcal{N}(0, \sigma^2 / 2)$.
\end{proof}

\begin{customprop}{\ref{prop:langevin}}
Let $\gamma = \frac{\sigma_{i-1}}{\sigma_i}$. For $\alpha = \epsilon\cdot  \frac{\sigma_i^2}{\sigma_L^2}$ (as in \cref{alg:anneal}), we have $\bfx_T \sim \mcal{N}(\bfzero, s^2_T I)$, where
\begin{align}
    \frac{s^2_T}{\sigma_i^2} =  \bigg( 1 - \frac{\epsilon}{\sigma_L^2} \bigg)^{2T}\Bigg( \gamma^2 - \frac{2\epsilon}{\sigma_L^2 - \sigma_L^2 \left(1 - \frac{\epsilon}{\sigma_L^2} \right)^2} \Bigg) + \frac{2\epsilon}{\sigma_L^2 - \sigma_L^2 \left( 1 - \frac{\epsilon}{\sigma_L^2} \right)^2}.\label{eqn:step_size_2}
\end{align}
\end{customprop}

\begin{proof}
First, the conditions we know are
\begin{gather*}
    \bfx_0 \sim p_{\sigma_{i-1}}(\bfx) = \mcal{N}(\bfzero, \sigma_{i-1}^2 I),\\
    \bfx_{t+1} \gets \bfx_{t} + \alpha \nabla_\bfx \log p_{\sigma_i}(\bfx_t) + \sqrt{2\alpha} \bfz_t = \bfx_t -\alpha \frac{\bfx_t}{\sigma_i^2} + \sqrt{2\alpha} \bfz_t,
\end{gather*}
where $\bfz_t \sim \mcal{N}(\bfzero, I)$. Therefore, the variance of $\bfx_t$ satisfies
\begin{align*}
    \operatorname{Var}[\bfx_t] = \begin{cases}
        \sigma_{i-1}^2 I \quad & \text{if $t = 0$}\\
        \bigg(1 - \frac{\alpha}{\sigma_i^2} \bigg)^2\operatorname{Var}[\bfx_{t-1}] + 2\alpha I \quad & \text{otherwise}.
    \end{cases}
\end{align*}
Now let $\mbf{v} \triangleq \frac{2\alpha}{1 - \left( 1 - \frac{\alpha}{\sigma_i^2} \right)^2} I$, we have
\begin{align*}
    \operatorname{Var}[\bfx_{t}] - \mbf{v} = \bigg( 1 - \frac{\alpha}{\sigma_i^2} \bigg)^2(\operatorname{Var}[\bfx_{t-1}] - \mbf{v}).
\end{align*}
Therefore,
\begin{align}
    &\operatorname{Var}[\bfx_T] - \mbf{v} = \bigg( 1 - \frac{\alpha}{\sigma_i^2} \bigg)^{2T}(\operatorname{Var}[\bfx_{0}] - \mbf{v})\notag\\
    \implies & \operatorname{Var}[\bfx_T] = \bigg( 1 - \frac{\alpha}{\sigma_i^2} \bigg)^{2T}(\operatorname{Var}[\bfx_{0}] - \mbf{v}) + \mbf{v}\notag\\
    \implies & s_T^2 = \bigg( 1 - \frac{\alpha}{\sigma_i^2} \bigg)^{2T}\bigg(\sigma_{i-1}^2 - \frac{2\alpha}{1 - \big( 1 - \frac{\alpha}{\sigma_i^2} \big)^2}\bigg) + \frac{2\alpha}{1 - \big( 1 - \frac{\alpha}{\sigma_i^2} \big)^2}.\label{eqn:langevin_tmp}
\end{align}
Substituting $\epsilon ~\sigma_i^2 / \sigma_L^2$ for $\alpha$ in \cref{eqn:langevin_tmp}, we immediately obtain \cref{eqn:step_size_2}.
\end{proof}

\section{Experimental details}\label{app:exp_details}
\subsection{Network architectures and hyperparameters}\label{app:arch}
The original NCSN in \cite{song2019generative} uses a network structure based on RefineNet~\cite{lin2017refinenet}---a classical architecture for semantic segmentation. There are three major modifications to the original RefineNet in NCSN: (i) adding an enhanced version of conditional instance normalization (designed in \cite{song2019generative} and named CondInstanceNorm++) for every convolutional layer; (ii) replacing max pooling with average pooling in RefineNet blocks; and (iii) using dilated convolutions in the ResNet backend of RefineNet. We use exactly the same architecture for NCSN experiments, but for NCSNv2 or any other architecture implementing \cref{rec:cond}, we apply the following modifications: (i) setting the number of classes in CondInstanceNorm++ to 1 (which we name as InstanceNorm++); (ii) changing average pooling back to max pooling; and (iii) removing all normalization layers in RefineNet blocks. Here (ii) and (iii) do not affect the results much, but they are included because we hope to minimize the number of unnecessary changes to the standard RefineNet architecture (the original RefineNet blocks in \cite{lin2017refinenet} use max pooling and have no normalization layers). We name a ResNet block (with InstanceNorm++ instead of BatchNorm) ``ResBlock'', and a RefineNet block ``RefineBlock''. When CondInstanceNorm++ is added, we name them ``CondResBlock'' and ``CondRefineBlock'' respectively. We use the ELU activation function~\cite{clevert2015fast} throughout all architectures.

To ensure sufficient capacity and receptive fields, the network structures for images of different resolutions have different numbers of layers and filters. We summarize the architectures in \cref{tab:ncsn} and \cref{tab:ncsnv2}.
\begin{table}[h]
    \centering
    \caption{The architectures of NCSN for images of various resolutions.}\label{tab:ncsn}
    \begin{subtable}[t]{0.3\linewidth}
    \centering
    \caption{NCSN $32^2$--$64^2$}\label{tab:ncsn_small_arch}
    \begin{tabular}{c}
    \toprule\toprule
    3x3 Conv2D, 128\\
    \midrule
    CondResBlock, 128\\
    \midrule
    CondResBlock, 128\\
    \midrule
    CondResBlock down, 256\\
    \midrule
    CondResBlock, 256\\
    \midrule
    CondResBlock down, 256\\ dilation 2\\
    \midrule
    CondResBlock, 256 \\ dilation 2\\
    \midrule
    CondResBlock down, 256\\ dilation 4\\
    \midrule
    CondResBlock, 256\\ dilation 4\\
    \midrule
    CondRefineBlock, 256\\
    \midrule
    CondRefineBlock, 256\\
    \midrule
    CondRefineBlock, 128\\
    \midrule
    CondRefineBlock, 128\\
    \midrule
    3x3 Conv2D, 3\\
    \bottomrule
    \end{tabular}
    \end{subtable}
    \hspace{3cm}
    \begin{subtable}[t]{0.3\linewidth}
    \centering
    \caption{NCSN $96^2$--$128^2$}
    \begin{tabular}{c}
    \toprule\toprule
    3x3 Conv2D, 128\\
    \midrule
    CondResBlock, 128\\
    \midrule
    CondResBlock, 128\\
    \midrule
    CondResBlock down, 256\\
    \midrule
    CondResBlock, 256\\
    \midrule
    CondResBlock down, 256 \\
    \midrule
    CondResBlock, 256 \\
    \midrule
    CondResBlock down, 512\\ dilation 2\\
    \midrule
    CondResBlock, 512\\ dilation 2\\
    \midrule
    CondResBlock down, 512\\ dilation 4\\
    \midrule
    CondResBlock, 512 \\ dilation 4\\
    \midrule
    CondRefineBlock, 512\\
    \midrule
    CondRefineBlock, 256\\
    \midrule
    CondRefineBlock, 256\\
    \midrule
    CondRefineBlock, 128\\
    \midrule
    CondRefineBlock, 128\\
    \midrule
    3x3 Conv2D, 3\\
    \bottomrule
    \end{tabular}
    \end{subtable}
\end{table}

\begin{table}[H]
    \centering
    \caption{The architectures of NCSNv2 for images of various resolutions.}\label{tab:ncsnv2}
    \begin{subtable}[t]{0.3\linewidth}
        \centering
        \caption{NCSNv2 $32^2$--$64^2$}\label{tab:ncsnv2_small_arch}
        \begin{tabular}{c}
        \toprule \toprule
        3x3 Conv2D, 128\\
        \midrule
        ResBlock, 128\\
        \midrule
        ResBlock, 128 \\
        \midrule
        ResBlock down, 256\\
        \midrule
        ResBlock, 256\\
        \midrule
        ResBlock down, 256\\ dilation 2\\
        \midrule
        ResBlock, 256\\ dilation 2\\
        \midrule
        ResBlock down, 256\\ dilation 4\\
        \midrule
        ResBlock, 256\\ dilation 4\\
        \midrule
        RefineBlock, 256\\
        \midrule
        RefineBlock, 256\\
        \midrule
        RefineBlock, 128\\
        \midrule
        RefineBlock, 128\\
        \midrule
        3x3 Conv2D, 3\\
        \bottomrule
        \end{tabular}
    \end{subtable}
    \hfill
    \begin{subtable}[t]{0.3\linewidth}
        \centering
        \caption{NCSNv2 $96^2$--$128^2$}
        \begin{tabular}{c}
        \toprule \toprule
        3x3 Conv2D, 128\\
        \midrule
        ResBlock, 128\\
        \midrule
        ResBlock, 128 \\
        \midrule
        ResBlock down, 256\\
        \midrule
        ResBlock, 256\\
        \midrule
        ResBlock down, 256\\
        \midrule
        ResBlock, 256\\
        \midrule
        ResBlock down, 512\\ dilation 2\\
        \midrule
        ResBlock, 512\\ dilation 2\\
        \midrule
        ResBlock down, 512 \\ dilation 4\\
        \midrule
        ResBlock, 512 \\ dilation 4\\
        \midrule
        RefineBlock, 512\\
        \midrule
        RefineBlock, 256\\
        \midrule
        RefineBlock, 256\\
        \midrule
        RefineBlock, 128\\
        \midrule
        RefineBlock, 128\\
        \midrule
        3x3 Conv2D, 3\\
        \bottomrule
        \end{tabular}
    \end{subtable}
    \hfill
    \begin{subtable}[t]{0.3\linewidth}
        \centering
        \caption{NCSNv2 $256^2$}
        \begin{tabular}{c}
        \toprule \toprule
        3x3 Conv2D, 128\\
        \midrule
        ResBlock, 128\\
        \midrule
        ResBlock, 128 \\
        \midrule
        ResBlock down, 256\\
        \midrule
        ResBlock, 256\\
        \midrule
        ResBlock down, 256\\
        \midrule
        ResBlock, 256\\
        \midrule
        ResBlock down, 256\\
        \midrule
        ResBlock, 256\\
        \midrule
        ResBlock down, 512\\ dilation 2\\
        \midrule
        ResBlock, 512\\ dilation 2\\
        \midrule
        ResBlock down, 512 \\ dilation 4\\
        \midrule
        ResBlock, 512 \\ dilation 4\\
        \midrule
        RefineBlock, 512\\
        \midrule
        RefineBlock, 256\\
        \midrule
        RefineBlock, 256\\
        \midrule
        RefineBlock, 256\\
        \midrule
        RefineBlock, 128\\
        \midrule
        RefineBlock, 128\\
        \midrule
        3x3 Conv2D, 3\\
        \bottomrule
        \end{tabular}
    \end{subtable}
\end{table}

We use the Adam optimizer~\cite{kingma2014adam} for all models. When \cref{rec:cond} is not in effect, we choose the learning rate $0.001$; otherwise we use a learning rate $0.0001$ to avoid loss explosion. We set the $\epsilon$ parameter of Adam to $10^{-3}$ for FFHQ and $10^{-8}$ otherwise. We provide other hyperparameters in \cref{tab:hyperparameters}, where $\sigma_1$, $L$, $T$, and $\epsilon$ of NCSNv2 are all chosen in accordance with our proposed techniques. When the number of training data is larger than 60000, we randomly sample 10000 of them and compute the maximum pairwise distance, which is set as $\sigma_1$ for NCSNv2.

\begin{table}[H]
    \vspace{-1.2em}
    \caption{Hyperparameters of NCSN/NCSNv2. The latter is configured according to \cref{rec:init_noise}--\ref{rec:langevin}. $\sigma_1$ and $L$ determine the set of noise levels. $T$ and $\epsilon$ are parameters of annealed Langevin dynamics.}\label{tab:hyperparameters}
    \begin{center}
    \begin{adjustbox}{max width=\linewidth}
    \begin{tabular}{cccccccc}
    \toprule
    Model & Dataset & $\sigma_1$ & $L$ & $T$ & $\epsilon$ & Batch size & Training iterations\\ 
    \midrule
    NCSN & CIFAR-10 $32^2$ & 1 & 10 &  100 & 2e-5 & 128 & 300k\\
    NCSN & CelebA $64^2$ & 1 & 10 &  100 & 2e-5 & 128 & 210k\\
    NCSN & LSUN church\_outdoor $96^2$ & 1 & 10 & 100 & 2e-5 & 128 & 200k\\
    NCSN & LSUN bedroom $128^2$ & 1 & 10 & 100 & 2e-5 & 64 & 150k\\
    NCSNv2 & CIFAR-10 $32^2$ & 50 & 232 & 5 & 6.2e-6 & 128 & 300k\\
    NCSNv2 & CelebA $64^2$ & 90 & 500 & 5 & 3.3e-6 & 128 & 210k\\
    NCSNv2 & LSUN church\_outdoor $96^2$ & 140 & 788 & 4 & 4.9e-6 & 128 & 200k\\
    NCSNv2 & LSUN bedroom/tower $128^2$ & 190 & 1086 & 3 & 1.8e-6 & 128 & 150k\\
    NCSNv2 & FFHQ $256^2$ & 348 & 2311 & 3 & 0.9e-7 & 32 & 80k\\
    \bottomrule
    \end{tabular}
    \end{adjustbox}
    \end{center}
\end{table}

\subsection{Additional settings}\label{app:settings}
\textbf{Datasets:} We use the following datasets in our experiments: CIFAR-10~\cite{krizhevsky2009learning}, CelebA~\cite{liu2015faceattributes}, LSUN~\cite{yu2015lsun}, and FFHQ~\cite{karras2019style}. CIFAR-10 contains 50000 training images and 10000 test images, all of resolution $32\times 32$. CelebA contains 162770 training images and 19962 test images with various resolutions. For preprocessing, we first center crop them to size $140\times 140$, and then resize them to $64\times 64$. We choose the church\_outdoor, bedroom and tower categories in the LSUN dataset. They contain 126227, 3033042, and 708264 training images respectively, and all have 300 validation images. For preprocessing, we first resize them so that the smallest dimension of images is $96$ (for church\_outdoor) or $128$ (for bedroom and tower), and then center crop them to equalize their lengths and heights. Finally, the FFHQ dataset consists of 70000 high-quality facial images at resolution $1024\times 1024$. We resize them to $256\times 256$ in our experiments. Because FFHQ does not have an official test dataset, we randomly select 63000 images for training and the remaining 7000 as the test dataset. In addition, we apply random horizontal flip as data augmentation in all cases.

\textbf{Metrics:} We use FID~\cite{heusel2017gans} and HYPE$_\infty$~\cite{zhou2019hype} scores for quantitative comparison of results. When computing FIDs on CIFAR-10 $32\times 32$, we measure the distance between the statistics of samples and training data. When computing FIDs on CelebA $64\times 64$, we follow the settings in \cite{song2019bridging} where the distance is measured between 10000 samples and the test dataset. We use the official website \href{https://hype.stanford.edu}{https://hype.stanford.edu} for computing HYPE$_\infty$ scores. Regarding model selection, we follow the settings in \cite{song2019generative}, where we compute FID scores on 1000 samples every 5000 training iterations and choose the checkpoint with the smallest FID for computing both full FID scores (with more samples from it) and the HYPE$_\infty$ scores.

\textbf{Training:} We use the Adam~\cite{kingma2014adam} optimizer with default hyperparameters. The learning rates and batch sizes are provided in \cref{app:arch} and \cref{tab:hyperparameters}. We observe that for images at resolution $128\times 128$ or $256\times 256$, training can be unstable when the loss is near convergence. We note, however, this is a well-known problem of the Adam optimizer, and can be mitigated by techniques such as AMSGrad~\cite{reddi2019convergence}. We trained all models on Nvidia Tesla V100 GPUs.

\textbf{Settings for \cref{sec:cond}:}
The loss curves in \cref{fig:loss_compare} are results of two settings: (i) \cref{rec:init_noise}, \ref{rec:noise_levels}, \ref{rec:langevin} and \ref{rec:ema} are in effect, but the model architecture is the same as the original NCSN (\ie, \cref{tab:ncsn_small_arch}); and (ii) all techniques are in effect, \ie, the model is the same as NCSNv2 depicted in \cref{tab:ncsnv2_small_arch}. We apply EMA with momentum 0.9 to smooth the curves in \cref{fig:loss_compare}. We observe that despite being simpler to implement, the new noise conditioning method proposed in \cref{rec:cond} performs as well as the original and arguably more complex one in \cite{song2019generative} in terms of the training loss. See the ablation studies in \cref{sec:experiment} and \cref{app:ablation} for additional results.

\textbf{Interpolation:} We can interpolate between two different samples from NCSN/NCSNv2 via interpolating the Gaussian random noise injected by annealed Langevin dynamics. Specifically, suppose we have a total of $L$ noise levels, and for each noise level we run $T$ steps of Langevin dynamics. Let $\{\bfz_{ij}\}_{1\leq i\leq L, 1\leq j\leq T} \triangleq \{\bfz_{11}, \bfz_{12}, \cdots, \bfz_{1T}, \bfz_{21}, \bfz_{22}, \cdots, \bfz_{2T}, \cdots, \bfz_{L1}, \bfz_{L2}, \cdots, \bfz_{LT}\}$ denote the set of all Gaussian noise used in this procedure, where $\bfz_{ij}$ is the noise injected at the $j$-th iteration of Langevin dynamics corresponding to the $i$-th noise level. Next, suppose we have two samples $\bfx^{(1)}$ and $\bfx^{(2)}$ with the same initialization $\bfx_0$, and denote the corresponding set of Gaussian noise as $\{\bfz^{(1)}_{ij}\}_{1\leq i \leq L, 1\leq j \leq T}$ and $\{\bfz^{(2)}_{ij}\}_{1\leq i \leq L, 1\leq j \leq T}$ respectively. We can generate $N$ interpolated samples between $\bfx^{(1)}$ and $\bfx^{(2)}$, where for the $k$-th interpolated sample we use Gaussian noise $\{\cos \big(\frac{ k\pi}{2(N+1)}\big) \bfz_{ij}^{(1)} + \sin \big(\frac{k \pi}{2(N+1)}\big) \bfz_{ij}^{(2)}\}_{1\leq i\leq L, 1\leq j \leq T}$ and initialization $\bfx_0$.

\section{Additional experimental results}\label{app:exp_results}
\subsection{Additional results without the denoising step}\label{app:no_denoise_results}

\begin{figure}%
    \centering
    \begin{subfigure}[b]{0.43\textwidth}
        \includegraphics[width=\textwidth]{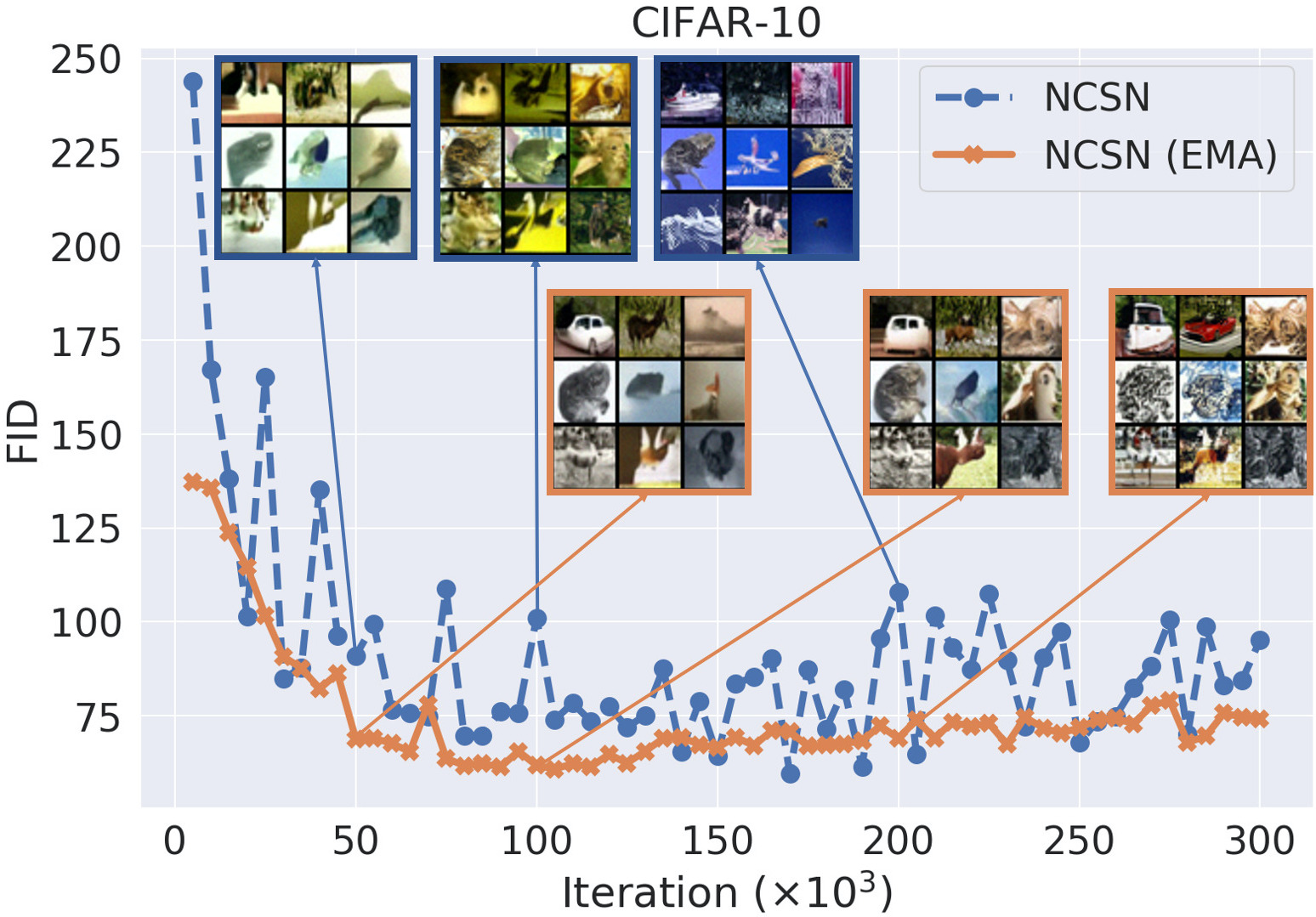}
    \end{subfigure}
    \begin{subfigure}[b]{0.43\textwidth}
        \includegraphics[width=\textwidth]{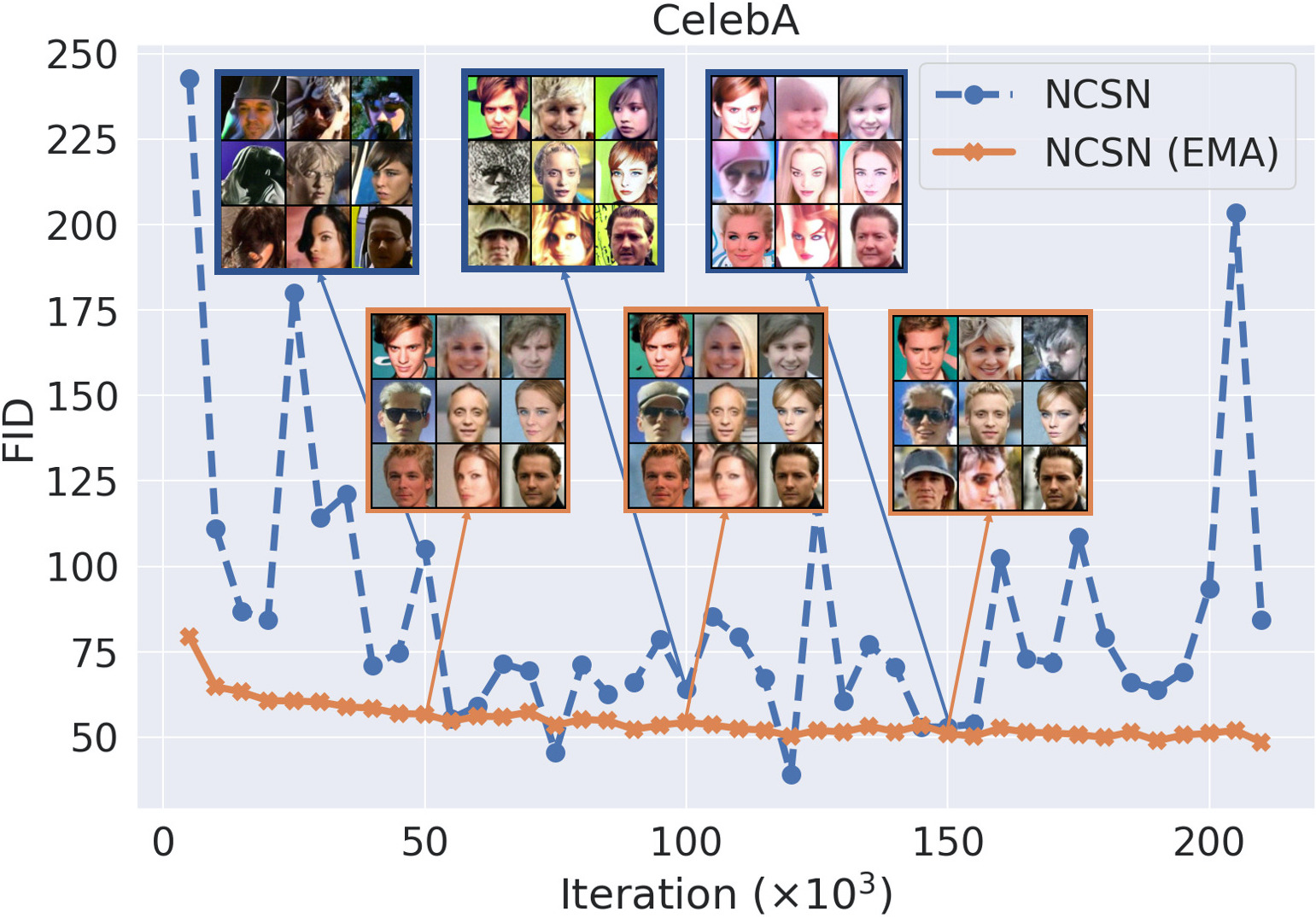}
    \end{subfigure}
    \caption{FIDs and color artifacts over the course of training (best viewed in color). The FIDs of NCSN have much higher volatility compared to NCSN with EMA. Samples from the vanilla NCSN often have obvious color shifts. All FIDs are computed without the denoising step.}
    \label{fig:stability_no_denosing}
\end{figure}

We further demonstrate the stabilizing effect of EMA in \cref{fig:stability_no_denosing}, where FIDs are computed without the denoising step. As indicated by \cref{fig:stability_no_denosing,fig:stability}, EMA can stabilize training and remove sample artifacts regardless of whether denoising is used or not.

\begin{figure}
    \centering
    \begin{minipage}{0.48\textwidth}
        \centering
            \begin{subfigure}[b]{0.5\textwidth}
            \includegraphics[width=\textwidth]{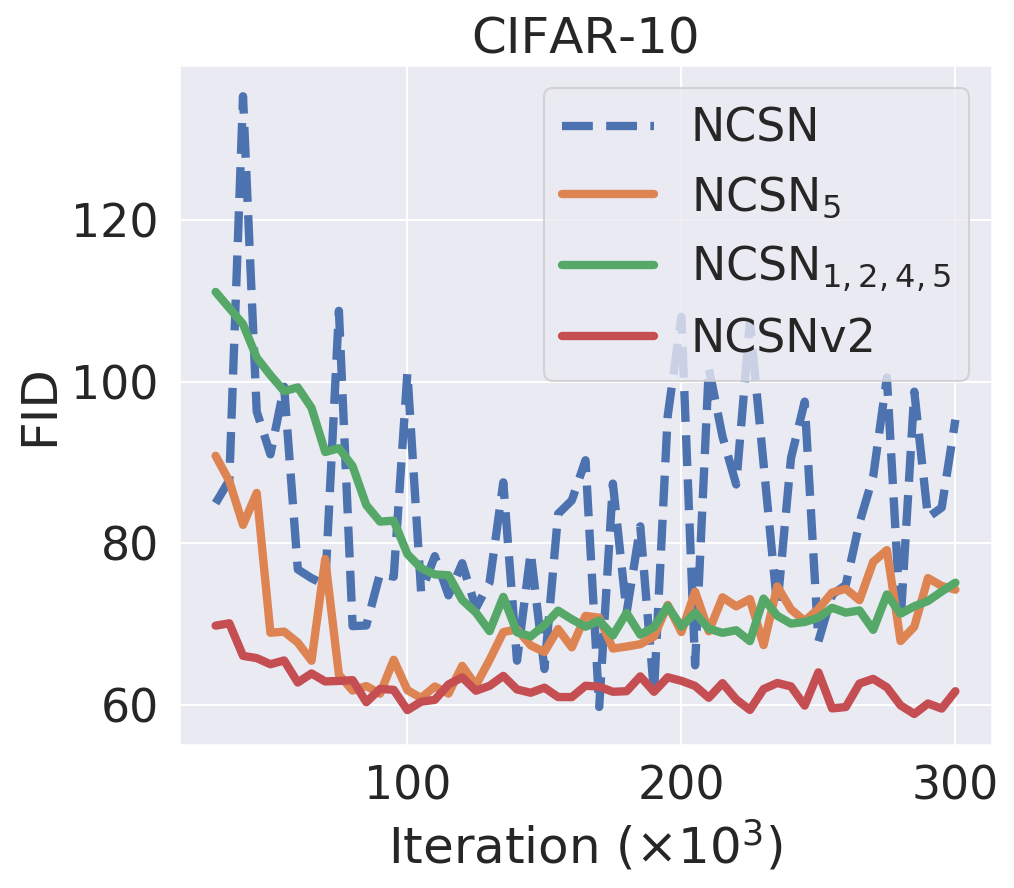}
            \caption{CIFAR-10 FIDs}\label{fig:fid_cifar10_no_denoising}
        \end{subfigure}%
        \begin{subfigure}[b]{0.5\textwidth}
            \includegraphics[width=\textwidth]{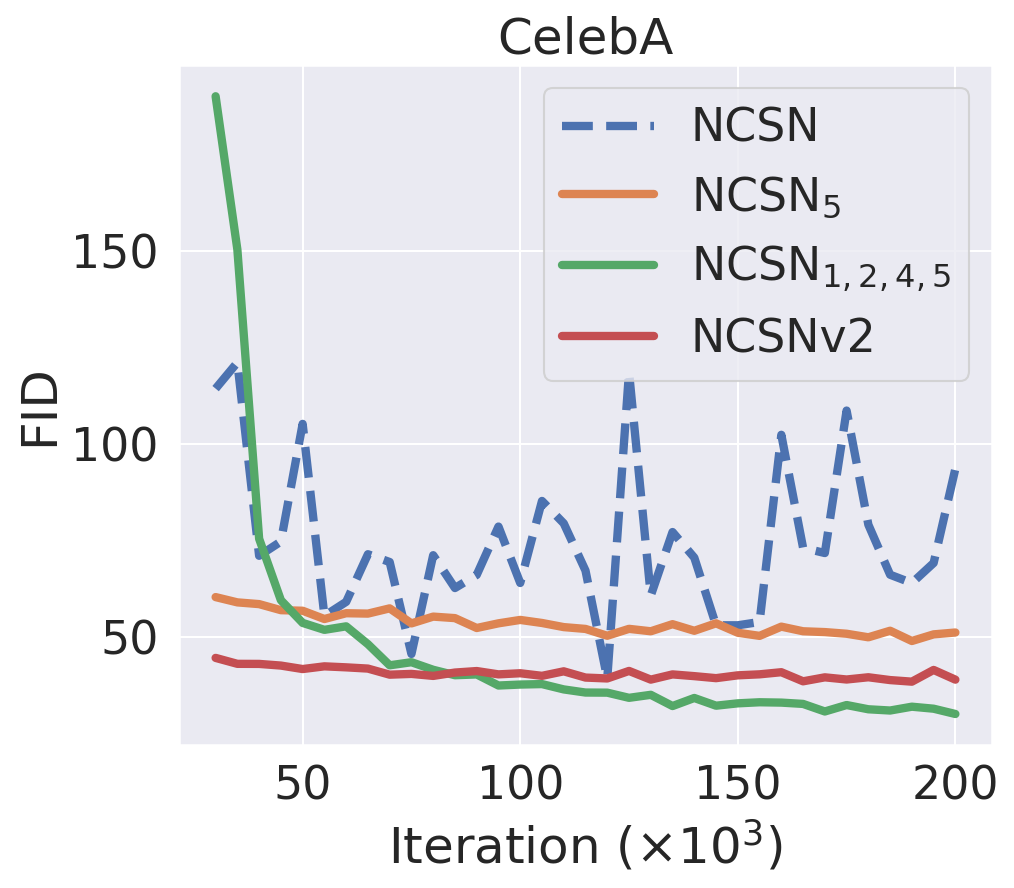}
            \caption{CelebA FIDs}\label{fig:fid_celeba_no_denoising}
        \end{subfigure}
        \caption{FIDs for different groups of techniques. Subscripts of ``NCSN'' are IDs of techniques in effect. ``NCSNv2'' uses all techniques. Results are computed without the denoising step.}
        \label{fig:ablation_no_denoising}
    \end{minipage}\hfill
    \begin{minipage}{0.45\textwidth}
        \centering
        \begin{subfigure}[b]{0.45\textwidth}
            \includegraphics[width=\textwidth]{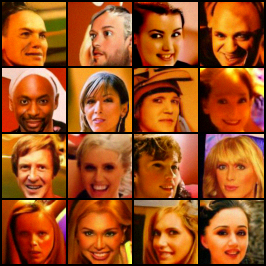}
            \caption{NCSN}\label{fig:celeba_ncsn_no_denoising}
        \end{subfigure}
        \begin{subfigure}[b]{0.45\textwidth}
            \includegraphics[width=\textwidth]{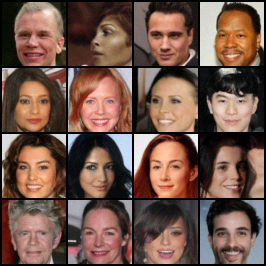}
            \caption{NCSNv2}\label{fig:celeba_ncsnv2_no_denoising}
        \end{subfigure}
        \caption{Uncurated samples from NCSN (a) and NCSNv2 (b) on CelebA $64\times 64$.}
    \end{minipage}
\end{figure}

FID scores should be interpreted with caution because they may not align well with human judgement. For example, the samples from NCSNv2 as demonstrated in \cref{fig:celeba_ncsnv2_no_denoising} have an FID score of 28.9 (without denoising), worse than NCSN (\cref{fig:celeba_ncsn_no_denoising}) whose FID is 26.9 (without denoising), but arguably produce much more visually appealing samples. To investigate whether FID scores align well with human ratings, we use the HYPE$_\infty$~\cite{zhou2019hype} score (higher is better), a metric of sample quality based on human evaluation, to compare the two models that generated samples in \cref{fig:celeba_ncsn_no_denoising,fig:celeba_ncsnv2_no_denoising}. We provide full results in \cref{tab:full_hype}, where all numbers except those for NCSN and NCSNv2 are directly taken from \cite{zhou2019hype}. As \cref{tab:full_hype} shows, our NCSNv2 achieves 37.3 on CelebA $64\times 64$ which is comparable to ProgressiveGAN~\cite{karras2017progressive}, whereas NCSN achieves 19.8. This is completely different from the ranking indicated by FIDs.

\begin{table}[H]
    \caption{HYPE$_\infty$ scores on CelebA $64\times 64$. $^\ast$With truncation tricks. }\label{tab:full_hype}
    \begin{center}
    \begin{tabular}{ccccc}
        \toprule
        Model & HYPE$_\infty$(\%) & Fakes Error(\%) & Reals Error(\%) & Std. \\ 
        \midrule
        StyleGAN$^\ast$~\cite{karras2019style} & 50.7 & 62.2 & 39.3 & 1.3\\
        ProgressiveGAN~\cite{karras2017progressive} & 40.3 & 46.2 & 34.4 & 0.9\\
        BEGAN~\cite{berthelot2017began} & 10 & 6.2 & 13.8 & 1.6\\
        WGAN-GP~\cite{gulrajani2017improved} & 3.8 & 1.7 & 5.9 & 0.6\\
        \midrule
        NCSN & 19.8 & 22.3 & 17.3 & 0.4\\
        NCSNv2 & 37.3 & 49.8 & 24.8 & 0.5\\
        \bottomrule\bigstrut
    {\footnotesize $^\ast$ with truncation tricks}
    \end{tabular}
    \end{center}
\end{table}

Finally, we provide ablation results without the denoising step in \cref{fig:ablation_no_denoising}. It is qualitatively similar to \cref{fig:ablation} where results are computed with denoising.

\subsection{Training and sampling speed}
In \cref{tab:speed}, we provide the time cost for training and sampling from NCSNv2 models on various datasets considered in our experiments.
\begin{table}[H]
    \caption{Training and sampling speed of NCSNv2 on various datasets.}\label{tab:speed}
    \begin{center}
	\begin{tabular}{cccc}
		\toprule
		Dataset & Device & Sampling time & Training time\\
		\midrule
		CIFAR-10 &	2x V100	& 2 min	& 22 h\\
		CelebA	& 4x V100 & 7 min & 29 h\\
		Church & 8x V100 & 17 min & 52 h\\
		Bedroom & 8x V100 & 19 min & 52 h\\
		Tower & 8x V100 & 19 min & 52 h\\
		FFHQ & 8x V100 & 50 min & 41 h\\
		\bottomrule
	\end{tabular}
	\end{center}
\end{table}

\subsection{Color shifts}\label{app:color}
\begin{figure}[H]
    \centering
    \begin{subfigure}{0.32\textwidth}
        \includegraphics[width=\textwidth]{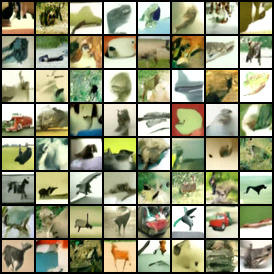}
        \caption{NCSN (Iter. = 50k)}
    \end{subfigure}
    \begin{subfigure}{0.32\textwidth}
        \includegraphics[width=\textwidth]{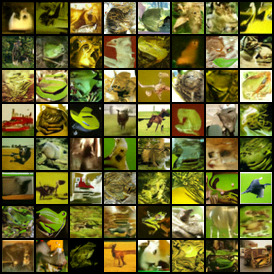}
        \caption{NCSN (Iter. = 100k)}
    \end{subfigure}
    \begin{subfigure}{0.32\textwidth}
        \includegraphics[width=\textwidth]{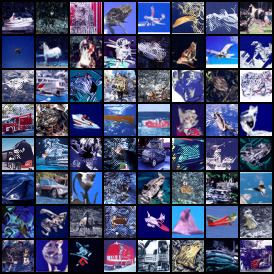}
        \caption{NCSN (Iter. = 200k)}
    \end{subfigure}\\
    \begin{subfigure}{0.32\textwidth}
        \includegraphics[width=\textwidth]{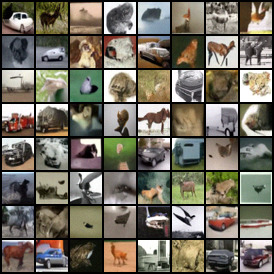}
        \caption{NCSN w/ EMA (Iter. = 50k)}
    \end{subfigure}
    \begin{subfigure}{0.32\textwidth}
        \includegraphics[width=\textwidth]{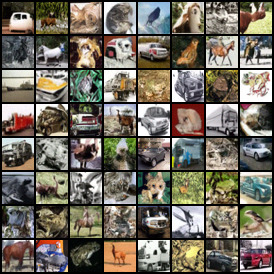}
        \caption{NCSN w/ EMA (Iter. = 100k)}
    \end{subfigure}
    \begin{subfigure}{0.32\textwidth}
        \includegraphics[width=\textwidth]{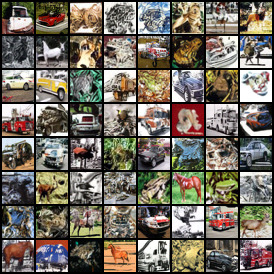}
        \caption{NCSN w/ EMA (Iter. = 200k)}
    \end{subfigure}
    \caption{EMA reduces undesirable color shifts on CIFAR-10. We show samples from NCSN and NCSN with EMA at the 50k/100k/200k-th training iteration.}
    \label{fig:color_shift_cifar10}
\end{figure}
\newpage
\vspace*{2.5cm}
\begin{figure}[H]
    \centering
    \begin{subfigure}{0.32\textwidth}
        \includegraphics[width=\textwidth]{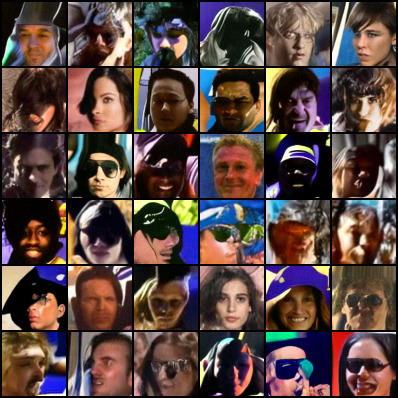}
        \caption{NCSN (Iter. = 50k)}
    \end{subfigure}
    \begin{subfigure}{0.32\textwidth}
        \includegraphics[width=\textwidth]{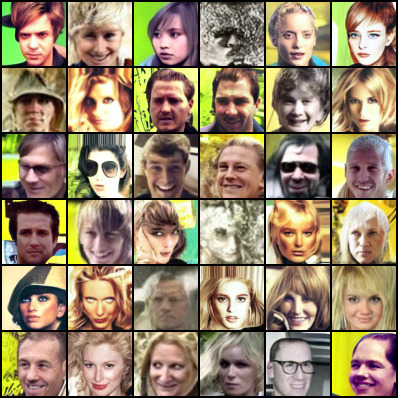}
        \caption{NCSN (Iter. = 100k)}
    \end{subfigure}
    \begin{subfigure}{0.32\textwidth}
        \includegraphics[width=\textwidth]{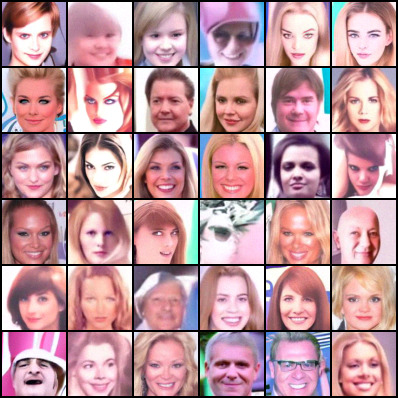}
        \caption{NCSN (Iter. = 150k)}
    \end{subfigure}\\
    \begin{subfigure}{0.32\textwidth}
        \includegraphics[width=\textwidth]{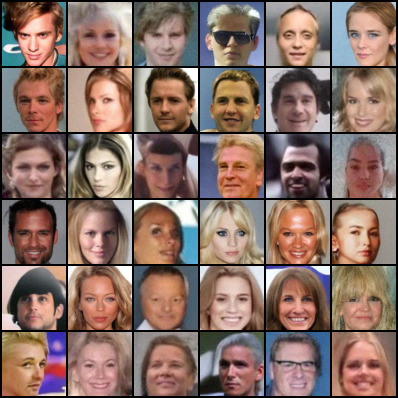}
        \caption{NCSN w/ EMA (Iter. = 50k)}
    \end{subfigure}
    \begin{subfigure}{0.32\textwidth}
        \includegraphics[width=\textwidth]{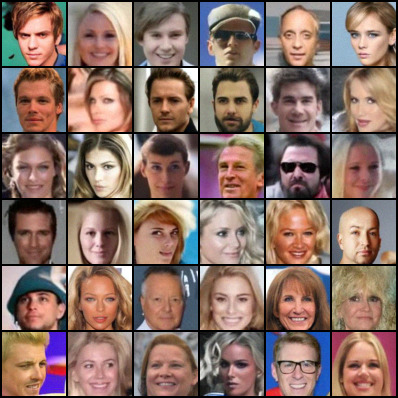}
        \caption{NCSN w/ EMA (Iter. = 100k)}
    \end{subfigure}
    \begin{subfigure}{0.32\textwidth}
        \includegraphics[width=\textwidth]{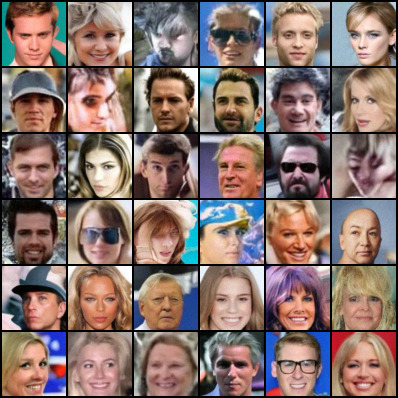}
        \caption{NCSN w/ EMA (Iter. = 150k)}
    \end{subfigure}
    \caption{EMA reduces undesirable color shifts on CelebA-10. We show samples from NCSN and NCSN with EMA at the 50k/100k/150k-th training iteration.}
    \label{fig:color_shift_celeba}
\end{figure}
\vspace{1cm}
\subsection{Additional results on ablation studies}\label{app:ablation}
\begin{figure}[H]
    \centering
    \begin{subfigure}{0.49\textwidth}
        \includegraphics[width=\textwidth]{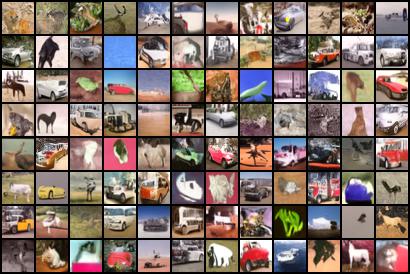}
        \caption{NCSN on CIFAR-10}
    \end{subfigure}
    \begin{subfigure}{0.49\textwidth}
        \includegraphics[width=\textwidth]{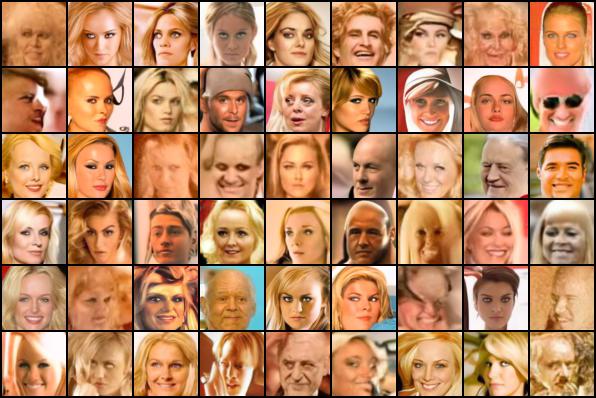}
        \caption{NCSN on CelebA}
    \end{subfigure}\\
    \begin{subfigure}{0.49\textwidth}
        \includegraphics[width=\textwidth]{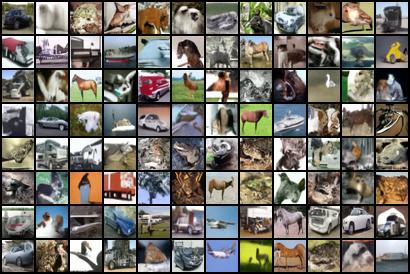}
        \caption{NCSN$_5$ on CIFAR-10}
    \end{subfigure}
    \begin{subfigure}{0.49\textwidth}
        \includegraphics[width=\textwidth]{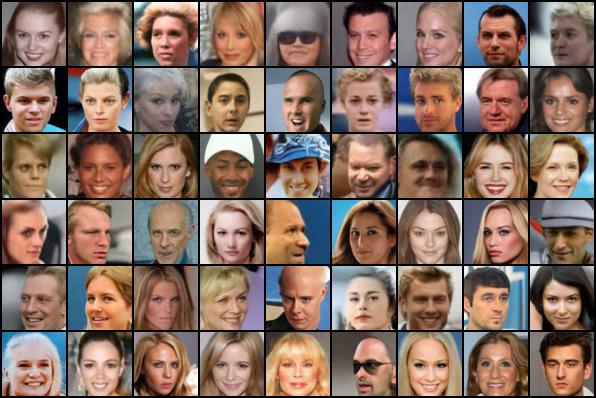}
        \caption{NCSN$_5$ on CelebA}
    \end{subfigure}\\
    \begin{subfigure}{0.49\textwidth}
        \includegraphics[width=\textwidth]{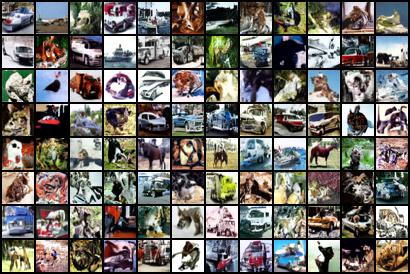}
        \caption{NCSN$_{1,2,4,5}$ on CIFAR-10}
    \end{subfigure}
    \begin{subfigure}{0.49\textwidth}
        \includegraphics[width=\textwidth]{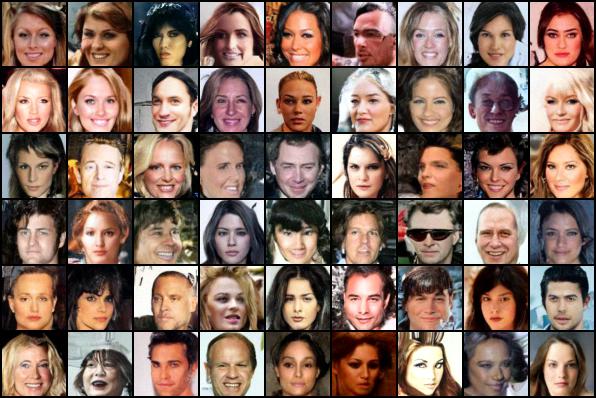}
        \caption{NCSN$_{1,2,4,5}$ on CelebA}
    \end{subfigure}\\
    \begin{subfigure}{0.49\textwidth}
        \includegraphics[width=\textwidth]{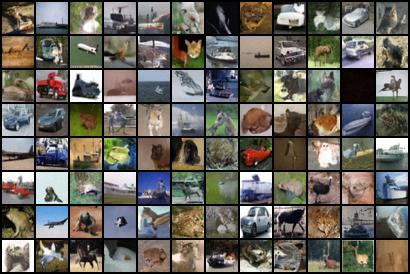}
        \caption{NCSNv2 on CIFAR-10}
    \end{subfigure}
    \begin{subfigure}{0.49\textwidth}
        \includegraphics[width=\textwidth]{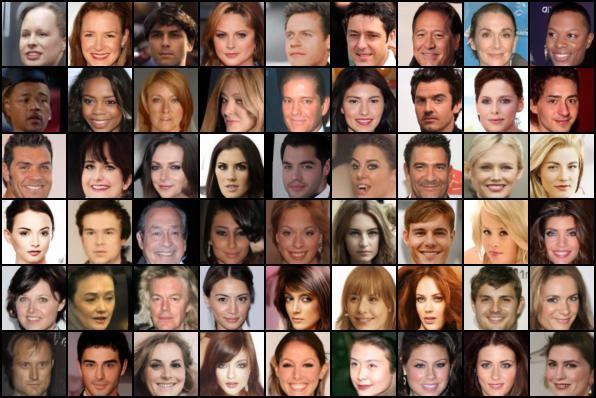}
        \caption{NCSNv2 on CelebA}
    \end{subfigure}
    \caption{Samples from models with different groups of techniques applied. NCSN is the original model in \cite{song2019generative} and does not use any of the newly proposed techniques. Subscripts of ``NCSN'' denote the IDs of techniques in effect. NCSN$_5$ only applies EMA. NCSN$_{1,2,4,5}$ applies both EMA and technique group (ii). NCSNv2 is the result of all techniques combined. Checkpoints are selected according to the lowest FID (with denoising) over the course of training.}
    \label{fig:ablation_samples}
\end{figure}

\begin{figure}[H]
    \centering
    \begin{subfigure}{0.49\textwidth}
        \includegraphics[width=\textwidth]{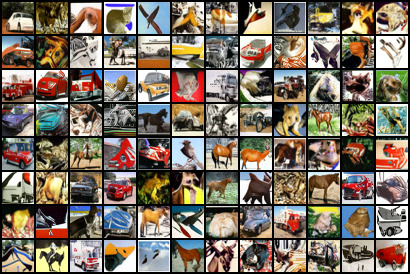}
        \caption{NCSN on CIFAR-10}
    \end{subfigure}
    \begin{subfigure}{0.49\textwidth}
        \includegraphics[width=\textwidth]{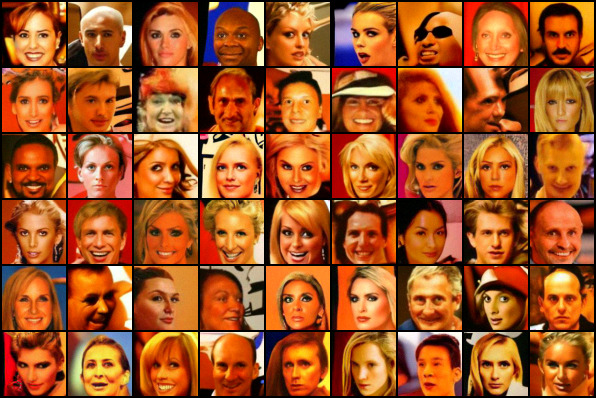}
        \caption{NCSN on CelebA}
    \end{subfigure}\\
    \begin{subfigure}{0.49\textwidth}
        \includegraphics[width=\textwidth]{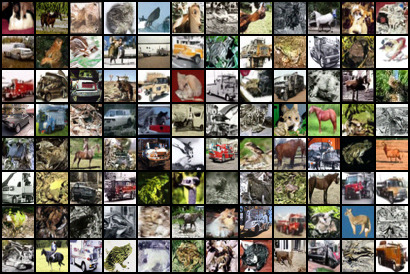}
        \caption{NCSN$_5$ on CIFAR-10}
    \end{subfigure}
    \begin{subfigure}{0.49\textwidth}
        \includegraphics[width=\textwidth]{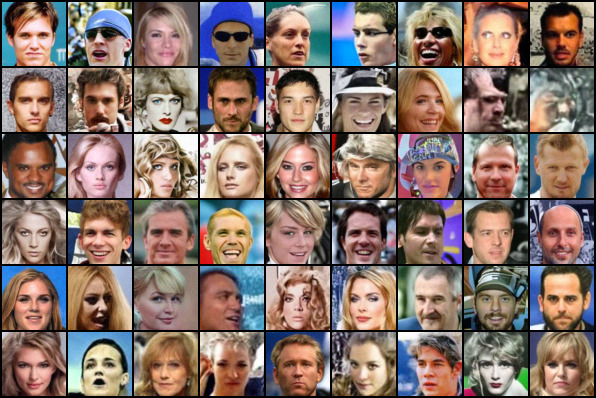}
        \caption{NCSN$_5$ on CelebA}
    \end{subfigure}\\
    \begin{subfigure}{0.49\textwidth}
        \includegraphics[width=\textwidth]{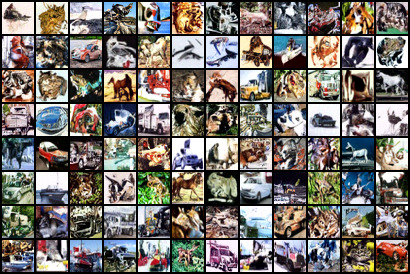}
        \caption{NCSN$_{1,2,4,5}$ on CIFAR-10}
    \end{subfigure}
    \begin{subfigure}{0.49\textwidth}
        \includegraphics[width=\textwidth]{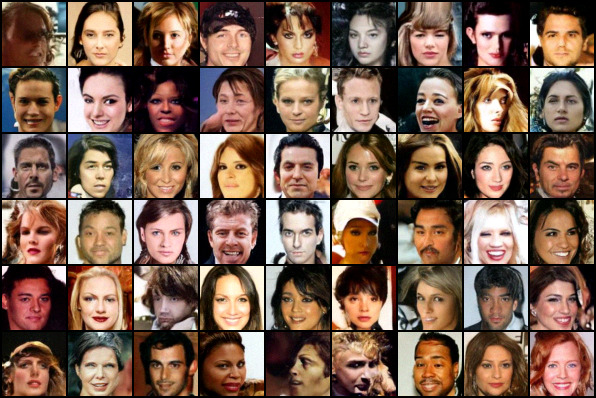}
        \caption{NCSN$_{1,2,4,5}$ on CelebA}
    \end{subfigure}\\
    \begin{subfigure}{0.49\textwidth}
        \includegraphics[width=\textwidth]{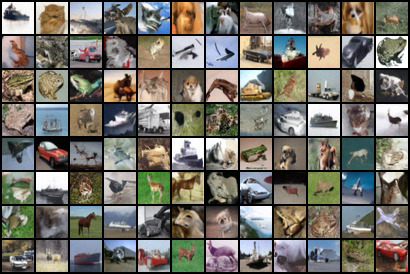}
        \caption{NCSNv2 on CIFAR-10}
    \end{subfigure}
    \begin{subfigure}{0.49\textwidth}
        \includegraphics[width=\textwidth]{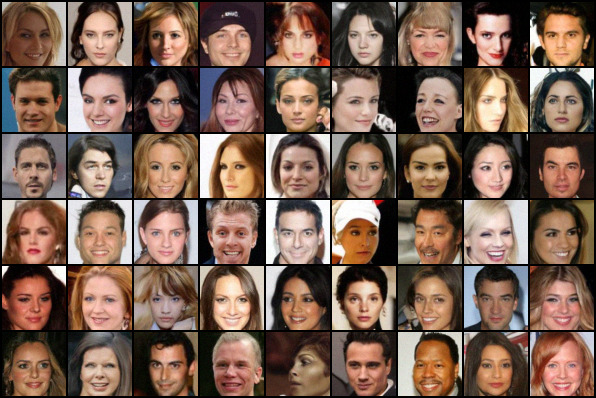}
        \caption{NCSNv2 on CelebA}
    \end{subfigure}
    \caption{Samples from models with different groups of techniques applied. NCSN is the original model in \cite{song2019generative} and does not use any of the newly proposed techniques. Subscripts of ``NCSN'' denote the IDs of techniques in effect. NCSN$_5$ only applies EMA. NCSN$_{1,2,4,5}$ applies both EMA and technique group (ii). NCSNv2 is the result of all techniques combined. Checkpoints are selected according to the lowest FID (without denoising) over the course of training.}
    \label{fig:ablation_samples_no_denoising}
\end{figure}

As discussed in \cref{sec:experiment}, we partition all techniques into three groups: (i) \cref{rec:ema}, (ii) \cref{rec:init_noise},\ref{rec:noise_levels},\ref{rec:langevin}, and (iii) \cref{rec:cond}, and investigate the performance of models after successively removing (iii), (ii), and (i) from NCSNv2. Aside from the FID curves in \cref{fig:ablation,fig:ablation_no_denoising}, we also provide samples from different models for visual inspection in \cref{fig:ablation_samples,fig:ablation_samples_no_denoising}. To generate these samples, we compute the FID scores on 1000 samples every 5000 training iterations for each considered model, and sample from the checkpoint of the smallest FID (the same setting as in \cite{song2019generative}). From samples in \cref{fig:ablation_samples,fig:ablation_samples_no_denoising}, we easily observe that removing any group of techniques leads to worse samples.

\newpage
\subsection{Generalization}\label{app:generalization}
\subsubsection{Loss curves}
\begin{figure}[H]
    \centering
    \begin{subfigure}{0.32\textwidth}
        \includegraphics[width=\textwidth]{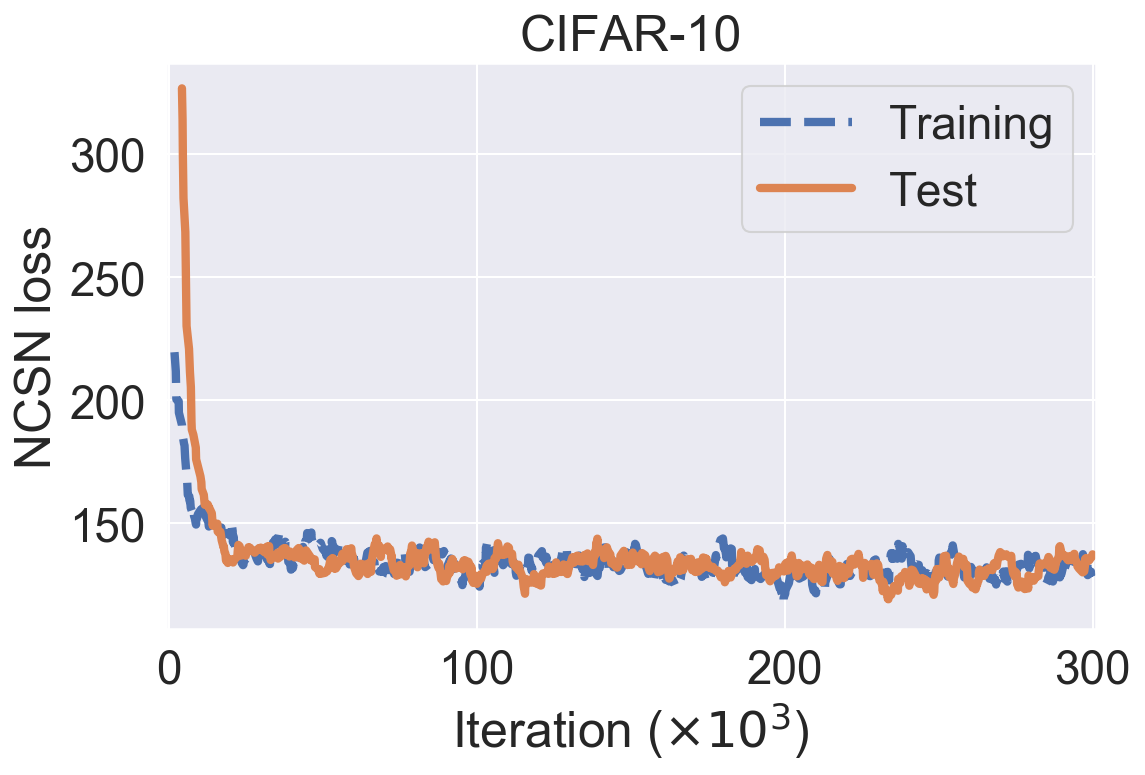}
        \caption{}
    \end{subfigure}
    \begin{subfigure}{0.32\textwidth}
        \includegraphics[width=\textwidth]{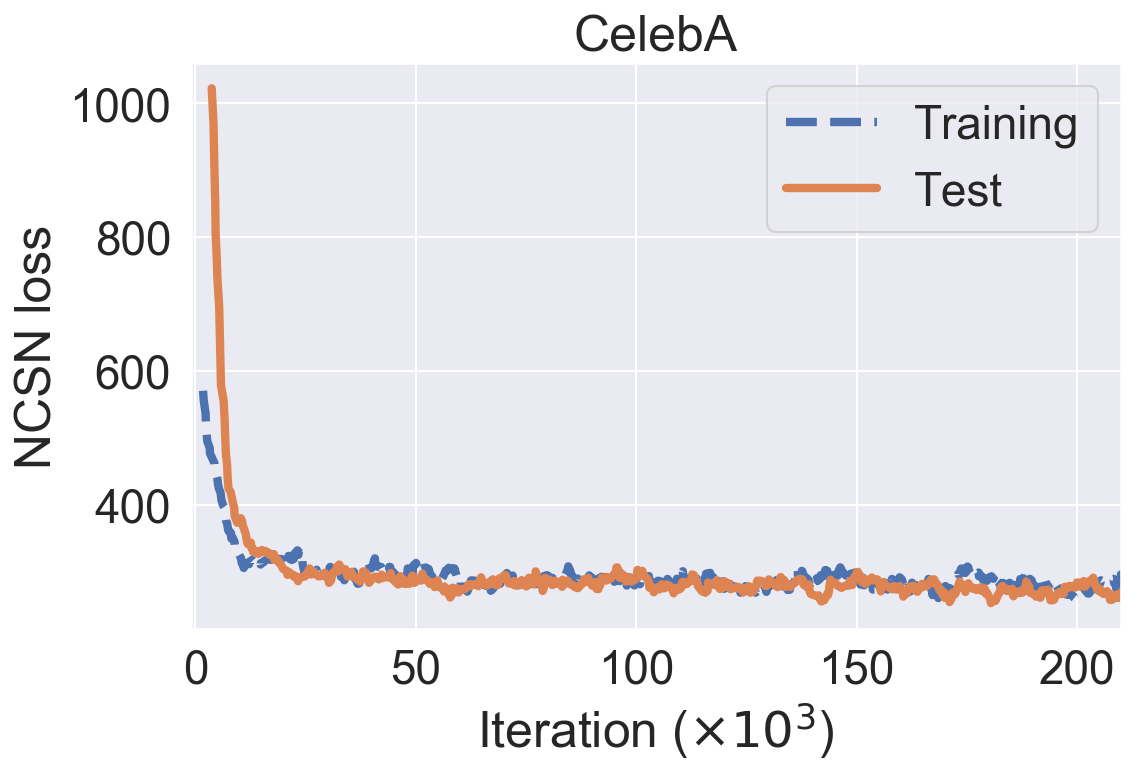}
        \caption{}
    \end{subfigure}
    \begin{subfigure}{0.32\textwidth}
        \includegraphics[width=\textwidth]{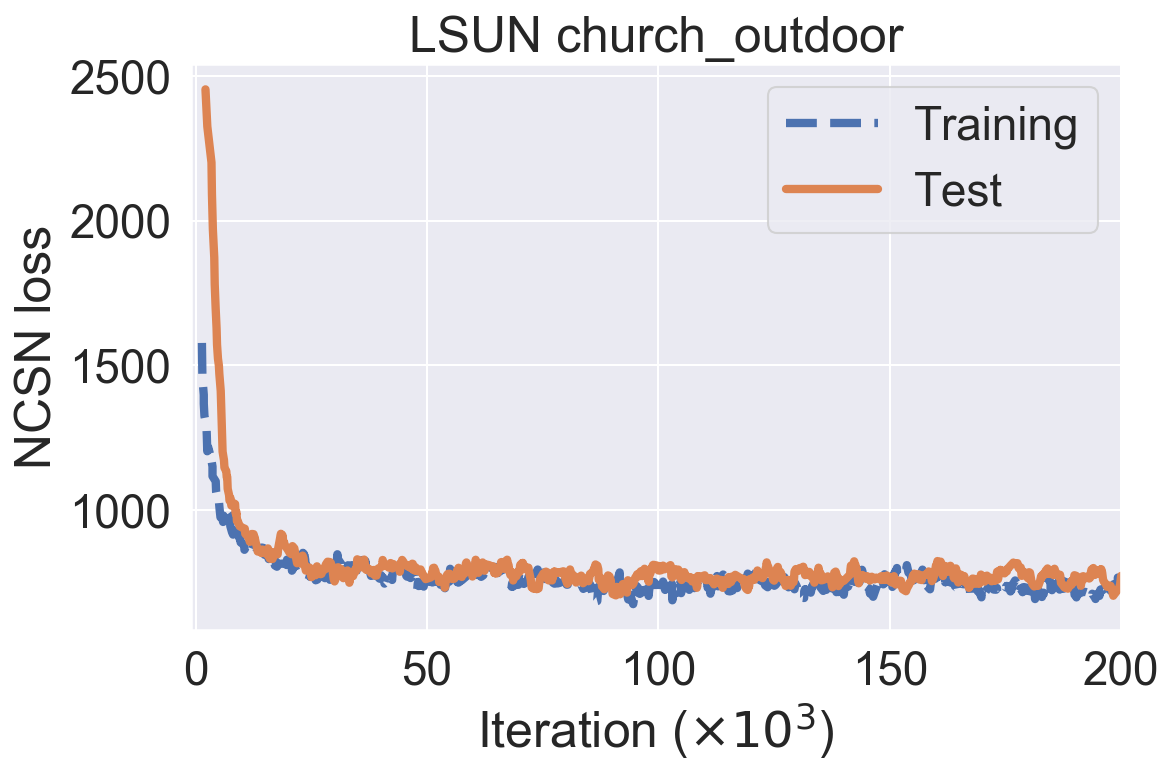}
        \caption{}
    \end{subfigure}\\
    \begin{subfigure}{0.32\textwidth}
        \includegraphics[width=\textwidth]{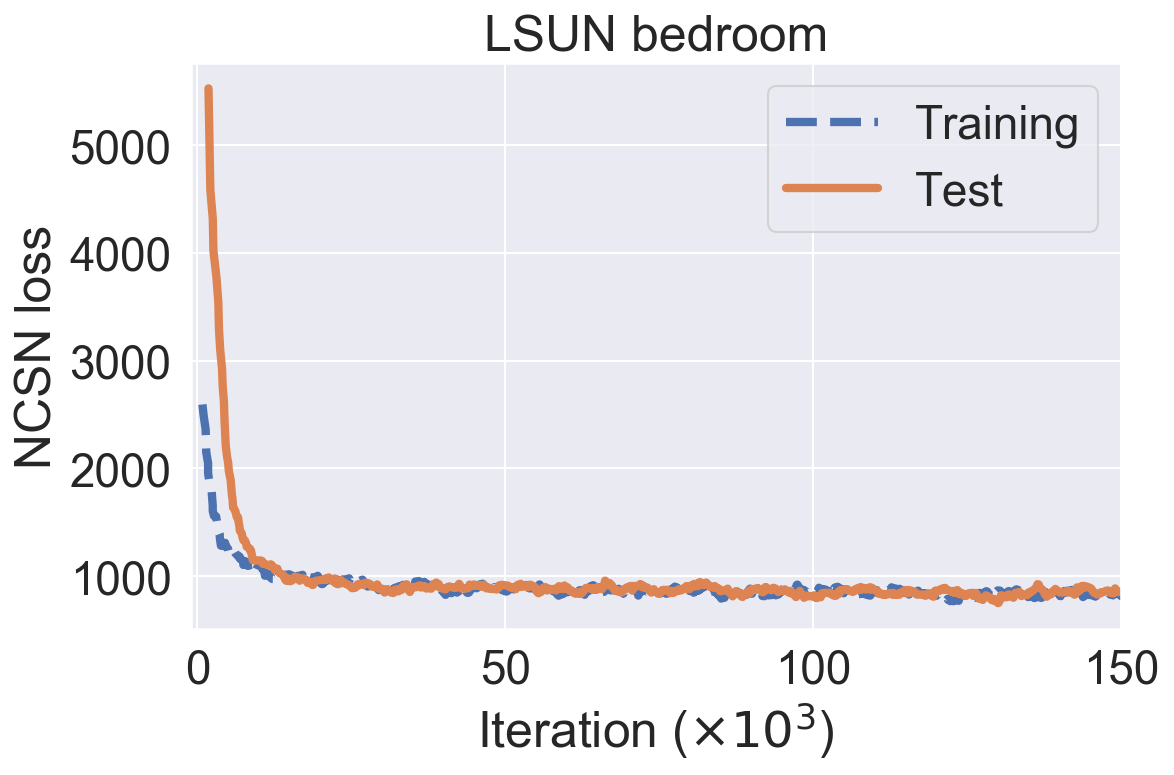}
        \caption{}
    \end{subfigure}
    \begin{subfigure}{0.32\textwidth}
        \includegraphics[width=\textwidth]{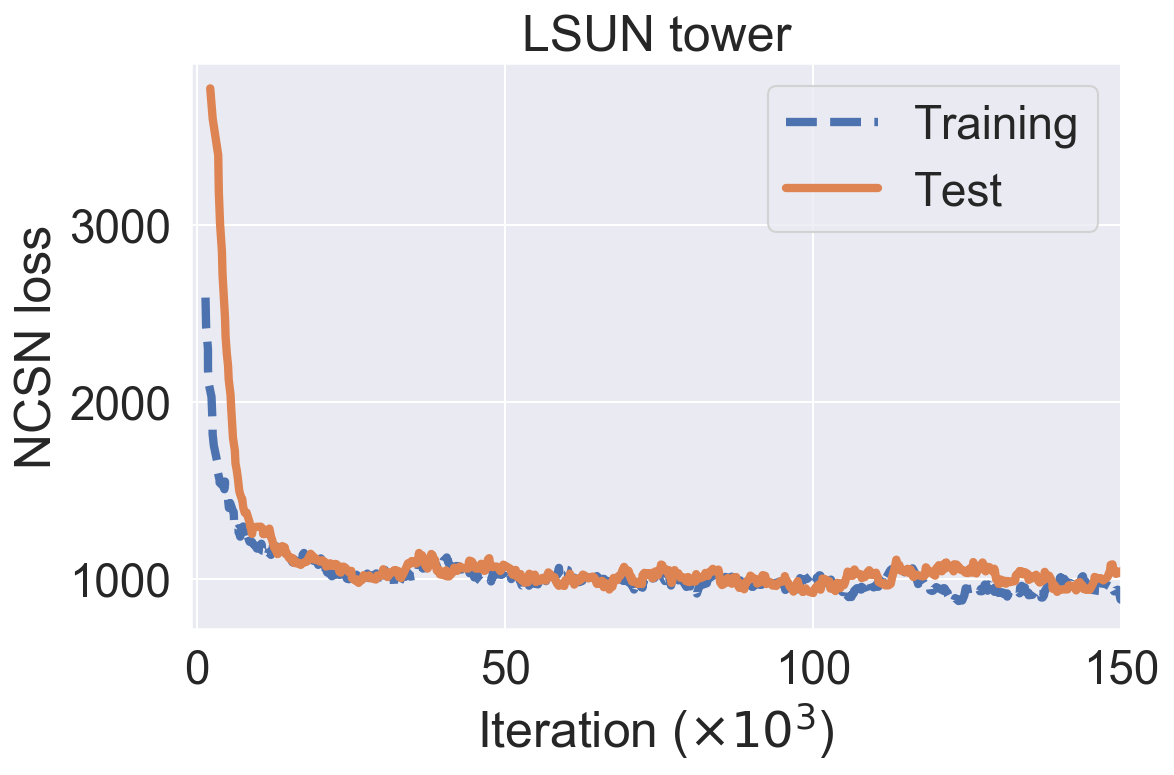}
        \caption{}
    \end{subfigure}
    \begin{subfigure}{0.32\textwidth}
        \includegraphics[width=\textwidth]{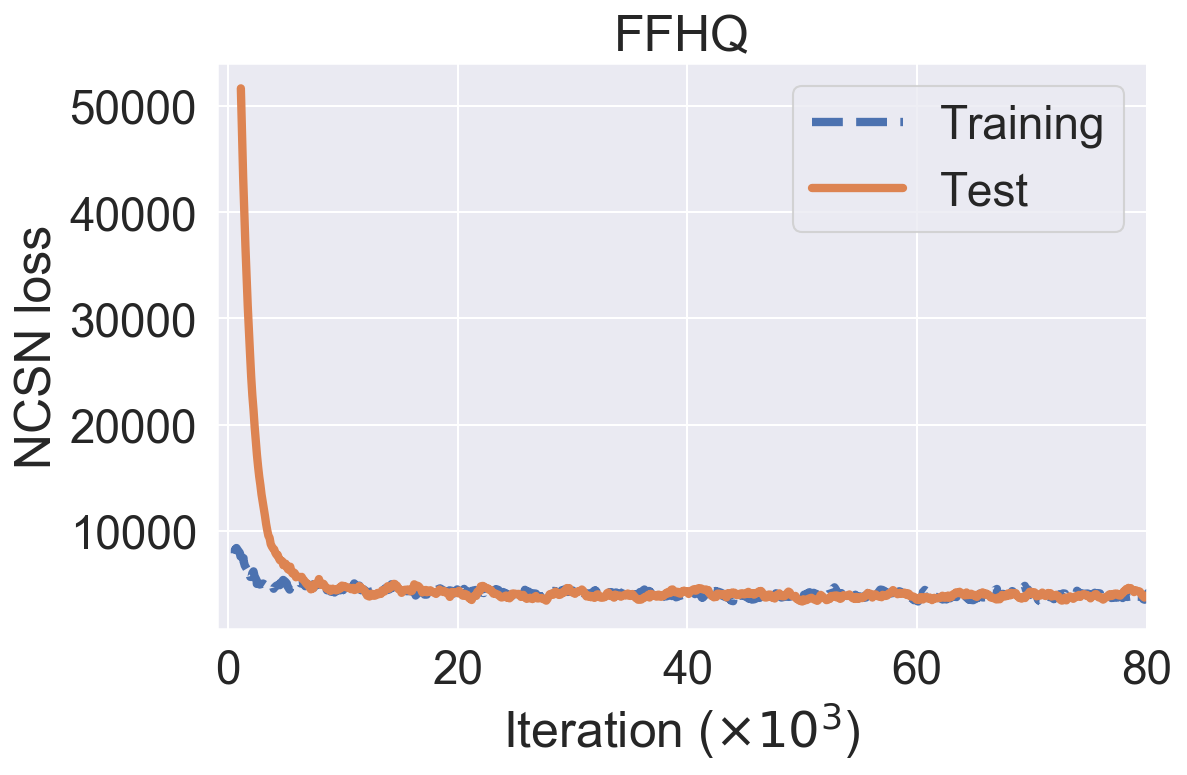}
        \caption{}
    \end{subfigure}
    \caption{Training vs. test loss curves of NCSNv2.}
    \label{fig:ncsnv2_train_test_curves}
\end{figure}

First, we demonstrate that our NCSNv2 does not overfit to the training dataset by showing the curves of training/test loss in \cref{fig:ncsnv2_train_test_curves}. Since the loss on the test dataset is always close to the loss on the training dataset during the course of training, this indicates that our model does not simply memorize training data.

\subsubsection{Nearest neighbors}
Starting from this section, all samples are from NCSNv2 at the last training iteration. For each generated sample, we show the nearest neighbors from the training dataset, measured by $\ell_2$ distance in the feature space of a pre-trained InceptionV3 network. Since we apply random horizontal flip when training, we also take this into consideration when computing nearest neighbors, so that we can detect cases in which NCSNv2 memorizes a flipped training data point.
\begin{figure}[H]
    \centering
    \includegraphics[width=\textwidth]{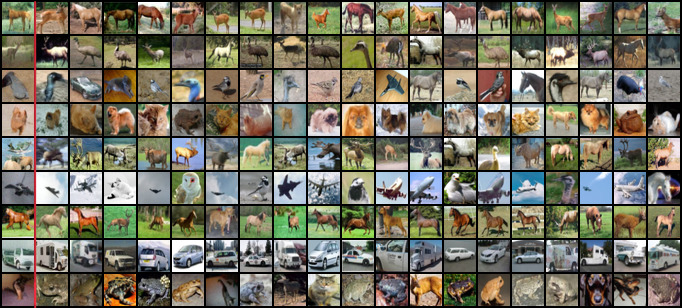}
    \caption{Nearest neighbors on CIFAR-10. NCSNv2 samples are on the left side of the red vertical line. Corresponding nearest neighbors are on the right side in the same row.}
    \label{fig:cifar10_nn}
\end{figure}

\begin{figure}[H]
    \centering
    \includegraphics[width=\textwidth]{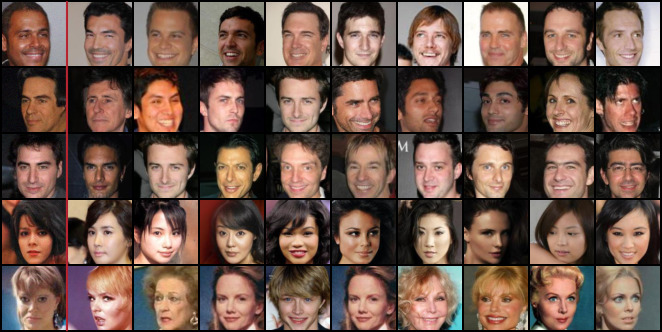}
    \caption{Nearest neighbors on CelebA $64\times 64$.}
    \label{fig:celeba_nn}
\end{figure}%
\begin{figure}[H]
    \centering
    \includegraphics[width=\textwidth]{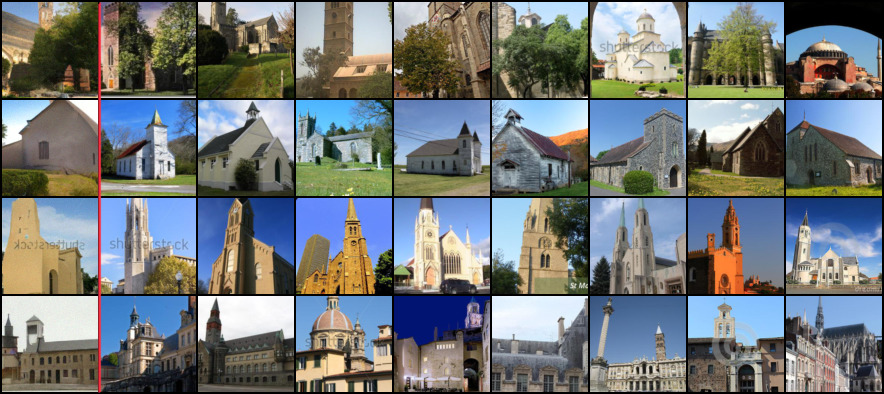}
    \caption{Nearest neighbors on LSUN church\_outdoor $96\times 96$.} %
    \label{fig:church_nn}
\end{figure}%
\begin{figure}[H]
    \centering
    \includegraphics[width=\textwidth]{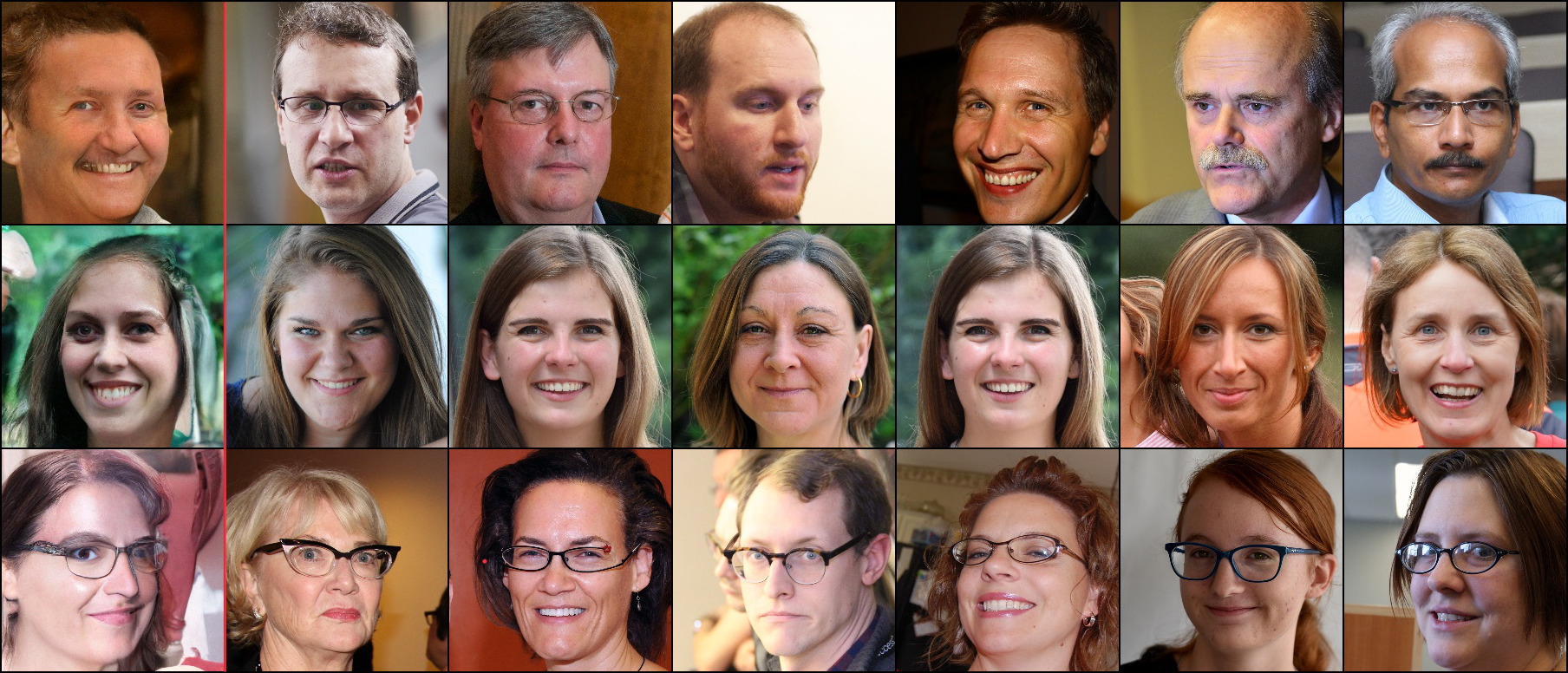}
    \caption{Nearest neighbors on FFHQ $256\times 256$.} %
    \label{fig:church_nn}
\end{figure}

\newpage
\subsubsection{Additional interpolation results}
We generate samples from NCSNv2 and interpolate between them using the method described in \cref{app:settings}. 
\begin{figure}[H]
    \centering
    \includegraphics[width=\textwidth]{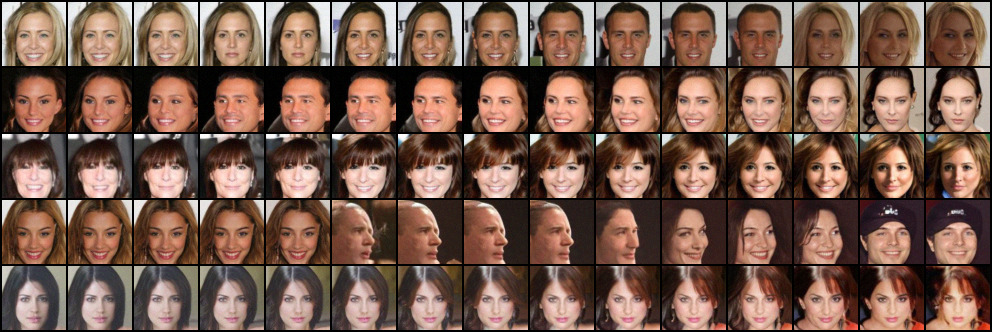}
    \caption{NCSNv2 interpolation results on CelebA $64\times 64$.}
\end{figure}
\begin{figure}[H]
    \centering
    \includegraphics[width=\textwidth]{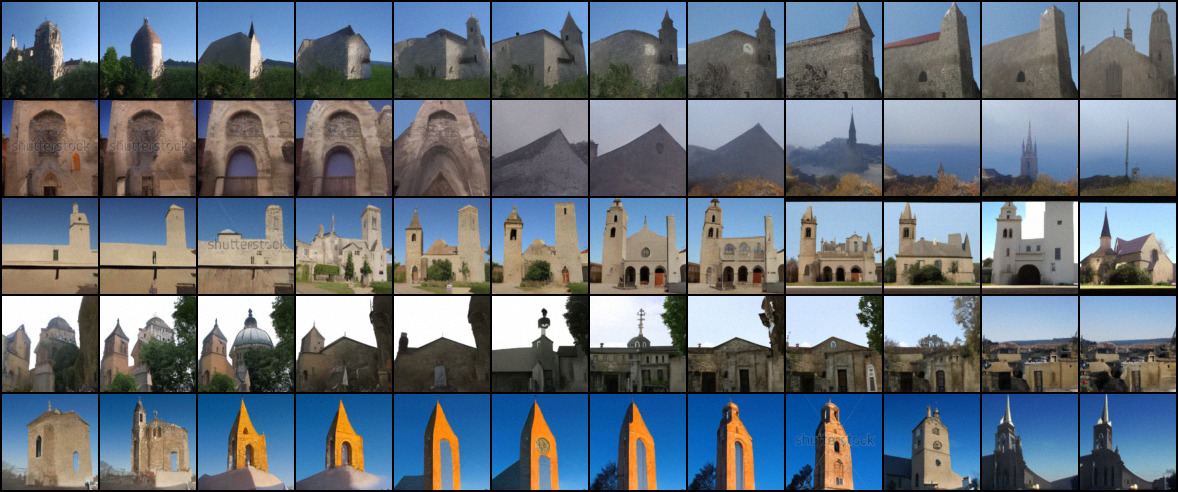}
    \caption{NCSNv2 interpolation results on LSUN church\_outdoor $96\times 96$.}
\end{figure}
\begin{figure}[H]
    \centering
    \includegraphics[width=\textwidth]{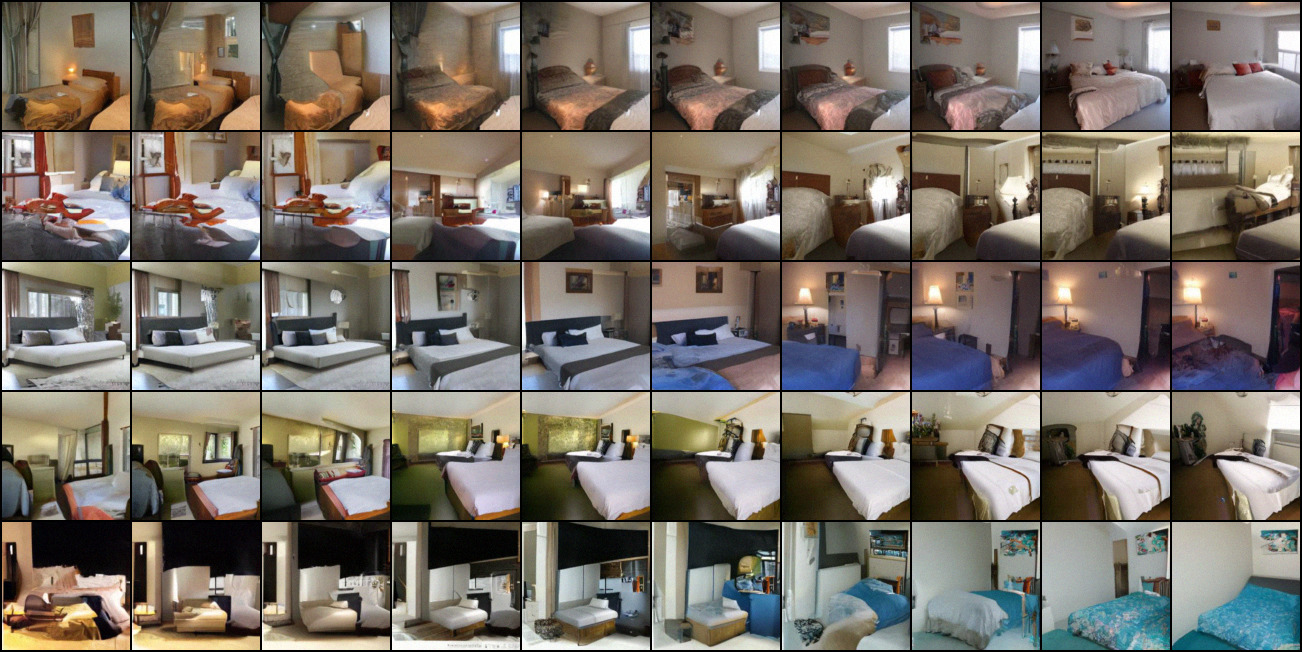}
    \caption{NCSNv2 interpolation results on LSUN bedroom $128\times 128$.}
\end{figure}
\newpage
\vspace*{\fill}
\begin{figure}[H]
    \centering
    \includegraphics[width=\textwidth]{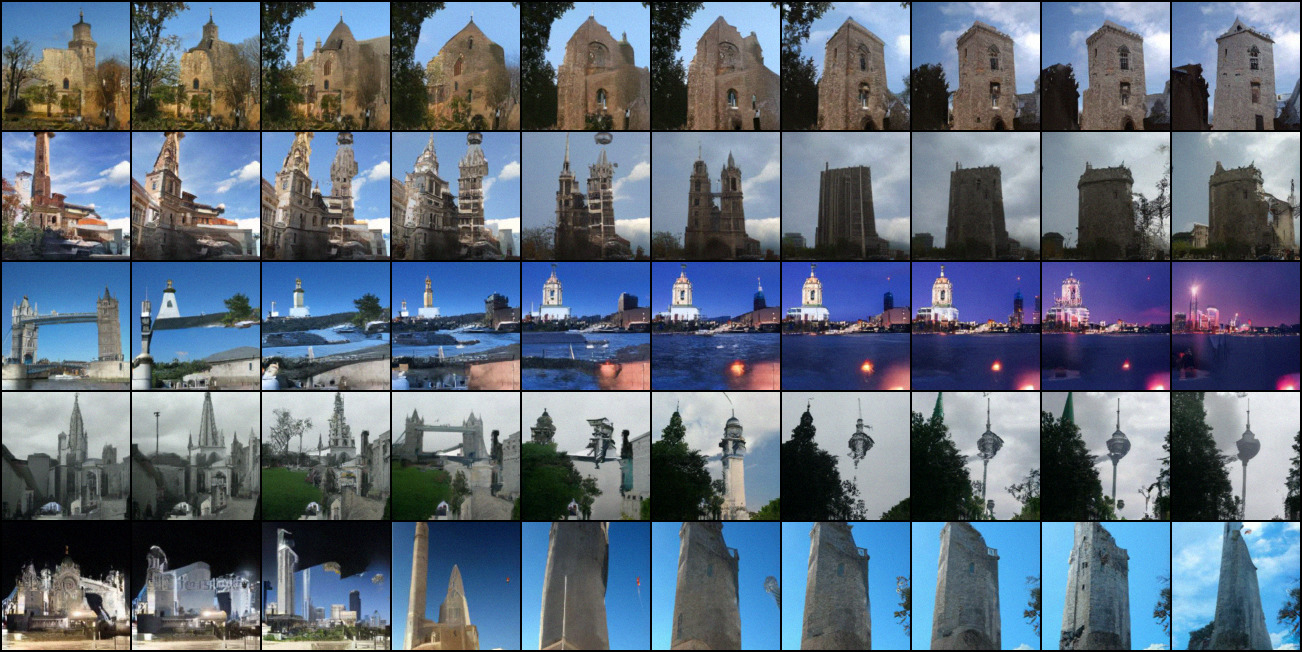}
    \caption{NCSNv2 interpolation results on LSUN tower $128\times 128$.}
\end{figure}
\begin{figure}[H]
    \centering
    \includegraphics[width=\textwidth]{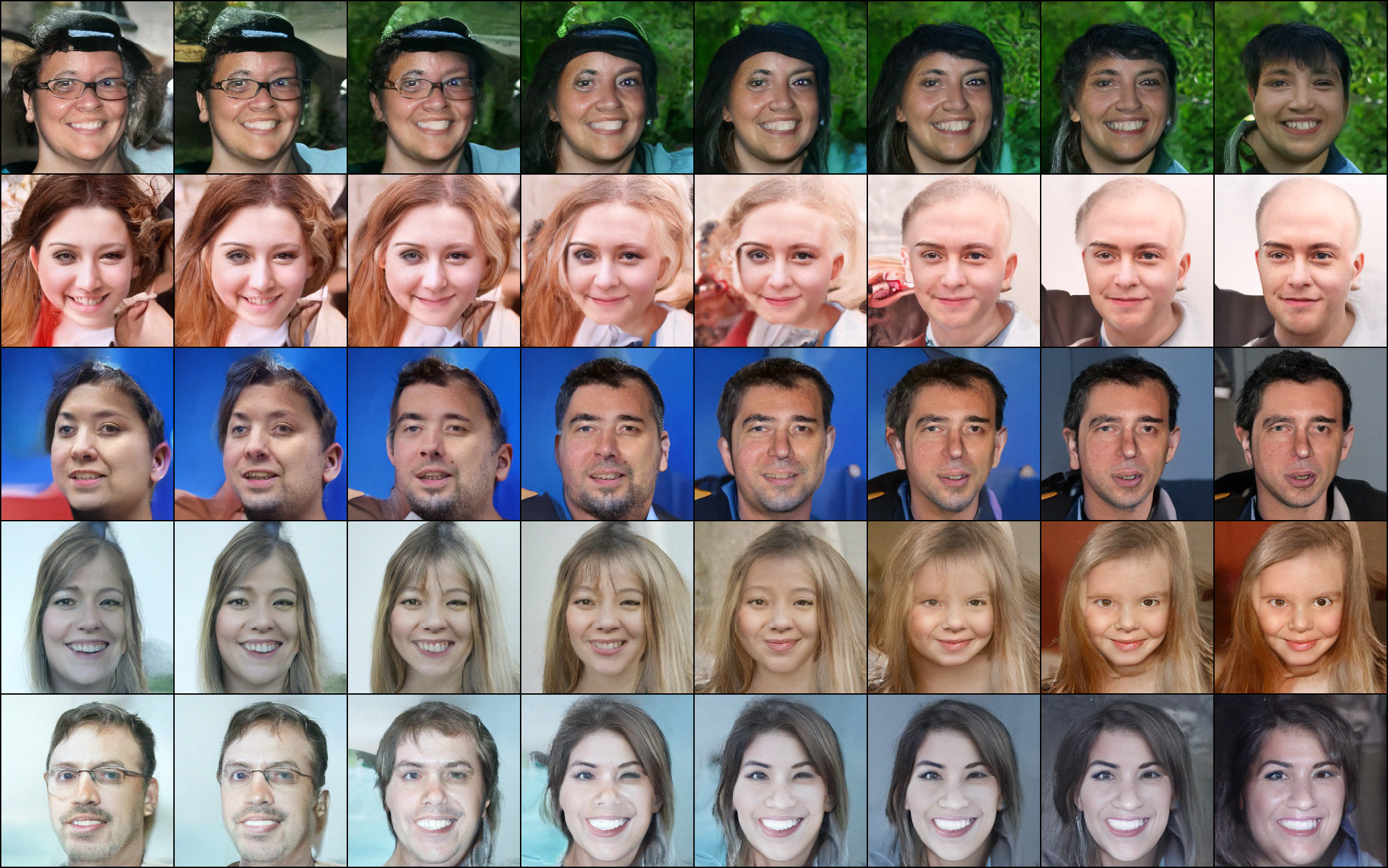}
    \caption{NCSNv2 interpolation results on FFHQ $256\times 256$.}
\end{figure}
\vspace*{\fill}

\newpage
\subsection{Additional uncurated samples}\label{app:samples}
\vspace*{\fill}
\begin{figure}[H]
    \centering
    \includegraphics[width=\textwidth]{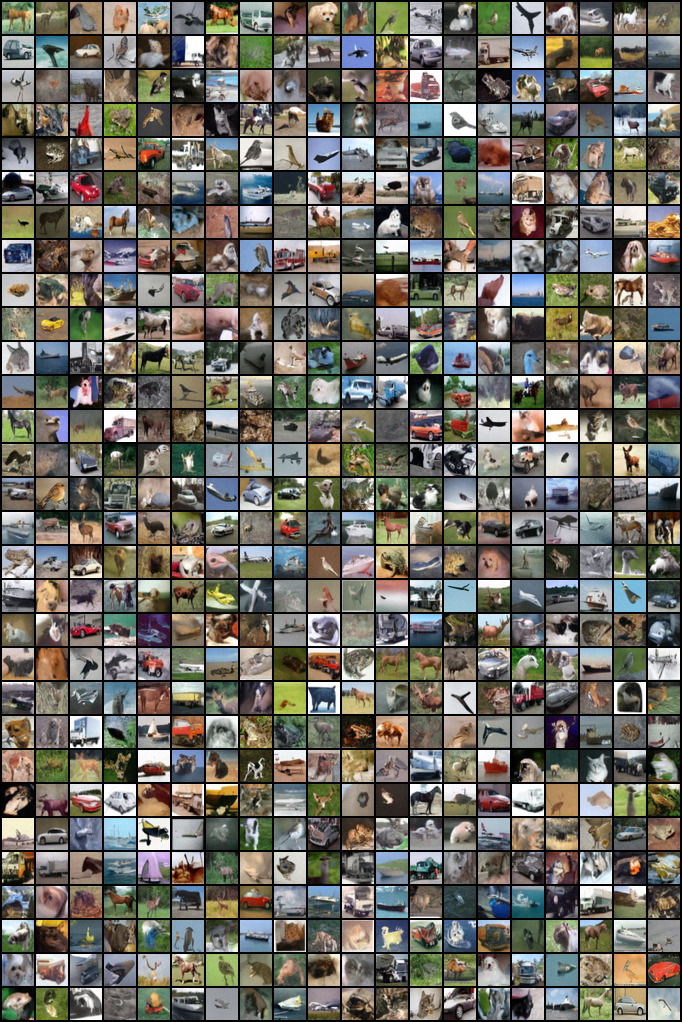}
    \caption{Uncurated CIFAR-10 $32\times 32$ samples from NCSNv2.}
    \label{fig:cifar10_full}
\end{figure}
\vspace*{\fill}

\newpage
\vspace*{\fill}
\begin{figure}[H]
    \centering
    \includegraphics[width=\textwidth]{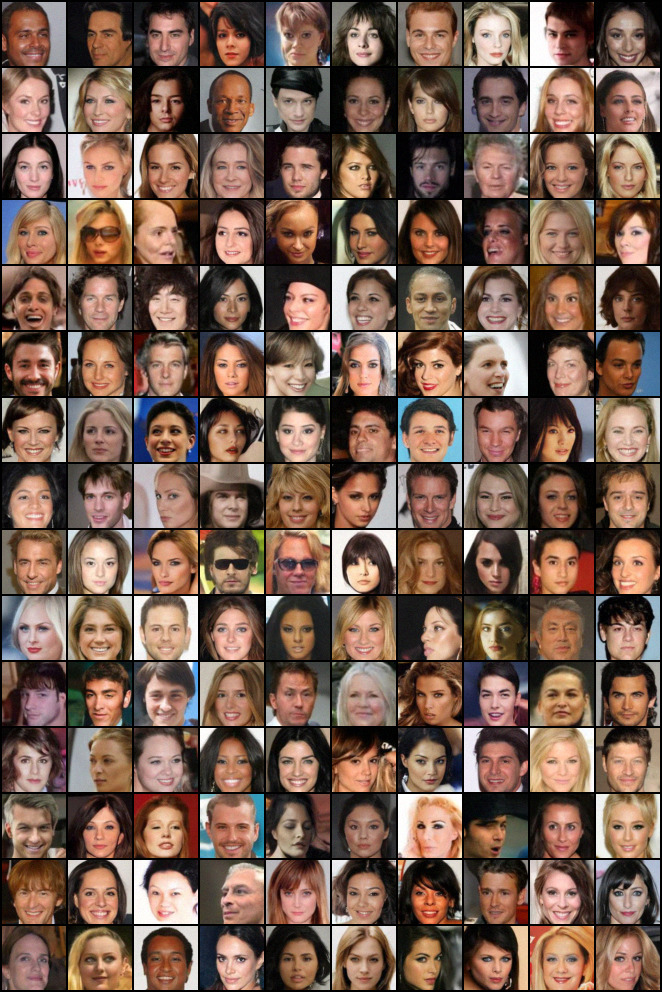}
    \caption{Uncurated CelebA $64\times 64$ samples from NCSNv2.}
    \label{fig:celeba_full}
\end{figure}
\vspace*{\fill}

\newpage
\vspace*{\fill}
\begin{figure}[H]
    \centering
    \includegraphics[width=\textwidth]{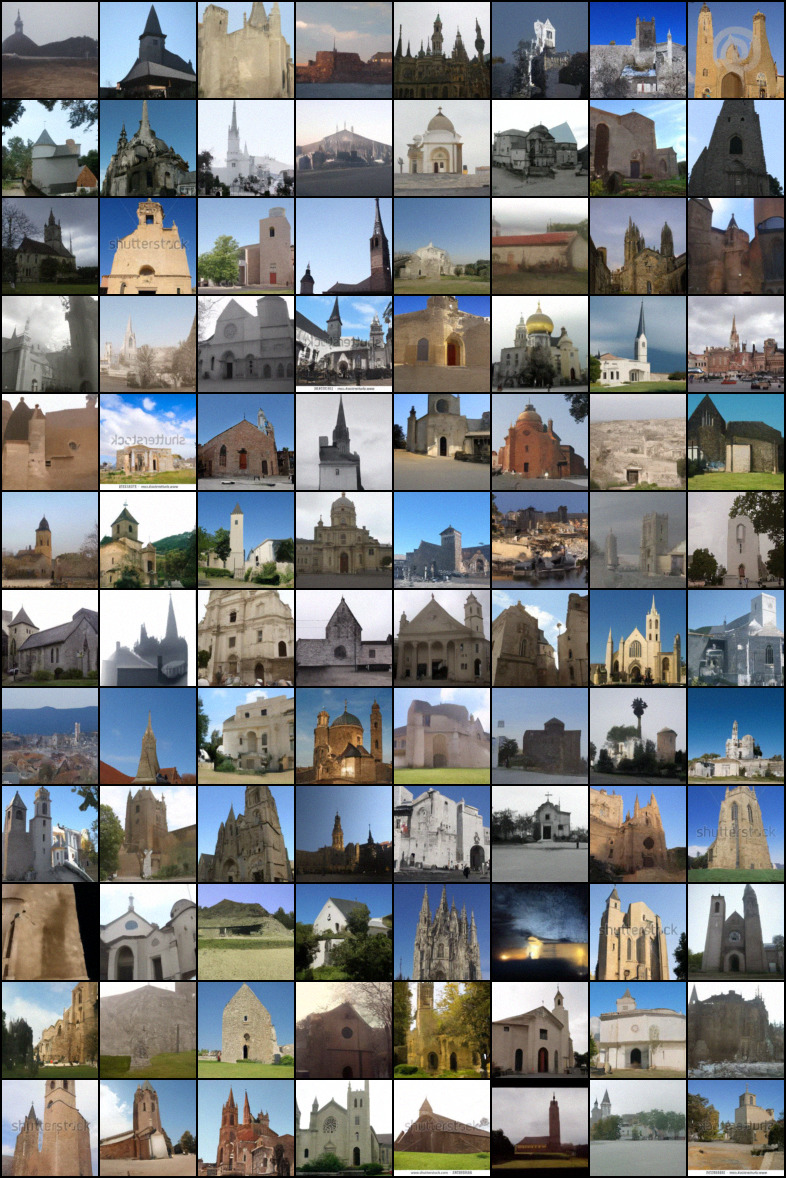}
    \caption{Uncurated LSUN church\_outdoor $96\times 96$ samples from NCSNv2.}
    \label{fig:church_full}
\end{figure}
\vspace*{\fill}

\newpage
\vspace*{\fill}
\begin{figure}[H]
    \centering
    \includegraphics[width=\textwidth]{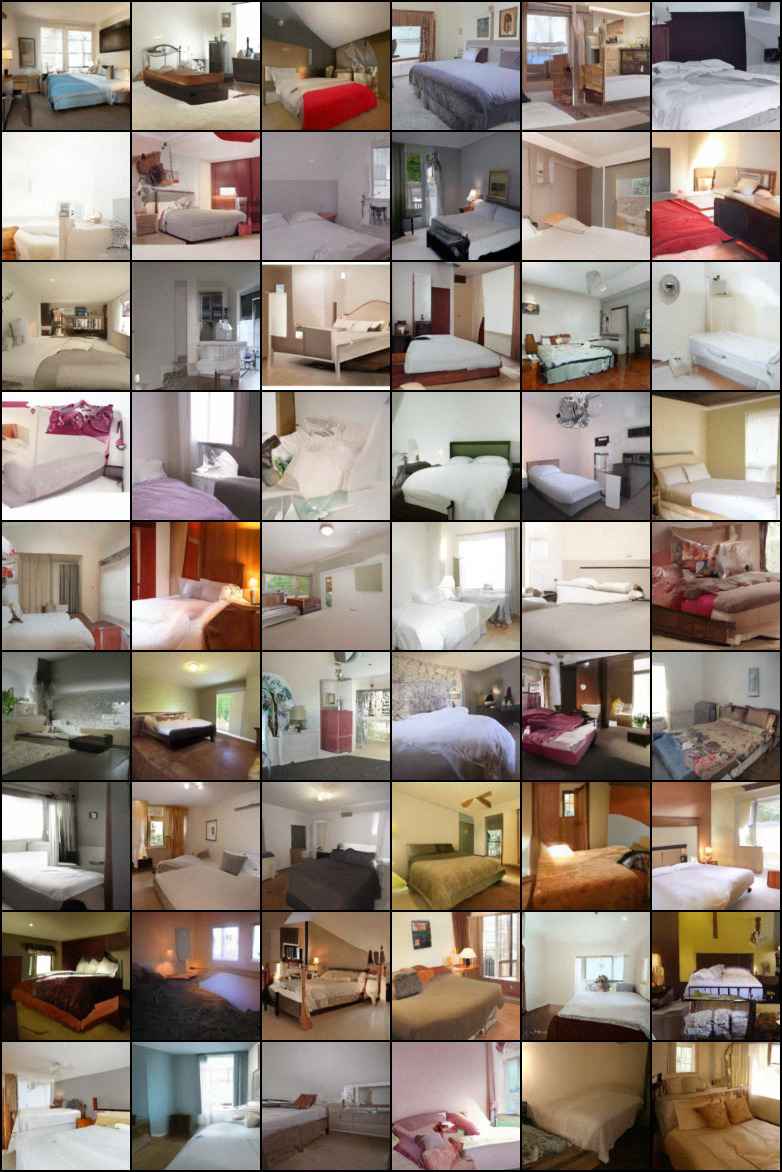}
    \caption{Uncurated LSUN bedroom $128\times 128$ samples from NCSNv2.}
    \label{fig:bedroom_full}
\end{figure}
\vspace*{\fill}

\newpage
\vspace*{\fill}
\begin{figure}[H]
    \centering
    \includegraphics[width=\textwidth]{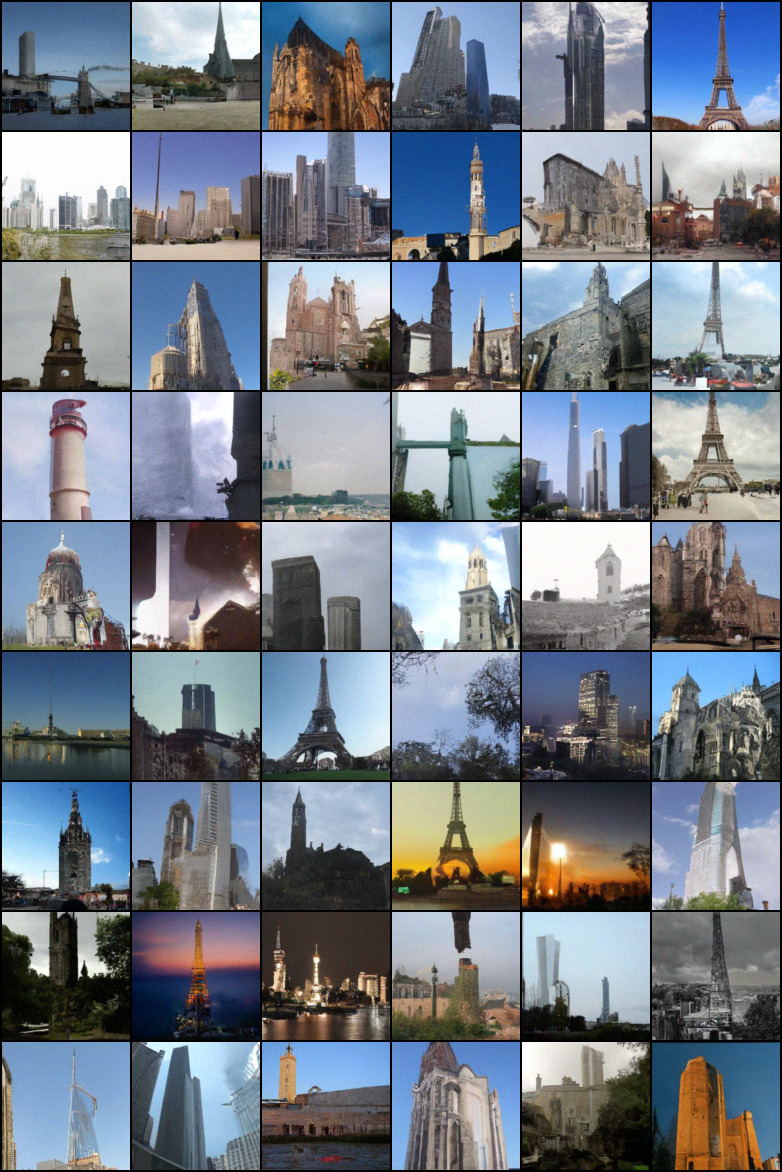}
    \caption{Uncurated LSUN tower $128\times 128$ samples from NCSNv2.}
    \label{fig:tower_full}
\end{figure}
\vspace*{\fill}

\newpage
\vspace*{\fill}
\begin{figure}[H]
    \centering
    \includegraphics[width=\textwidth]{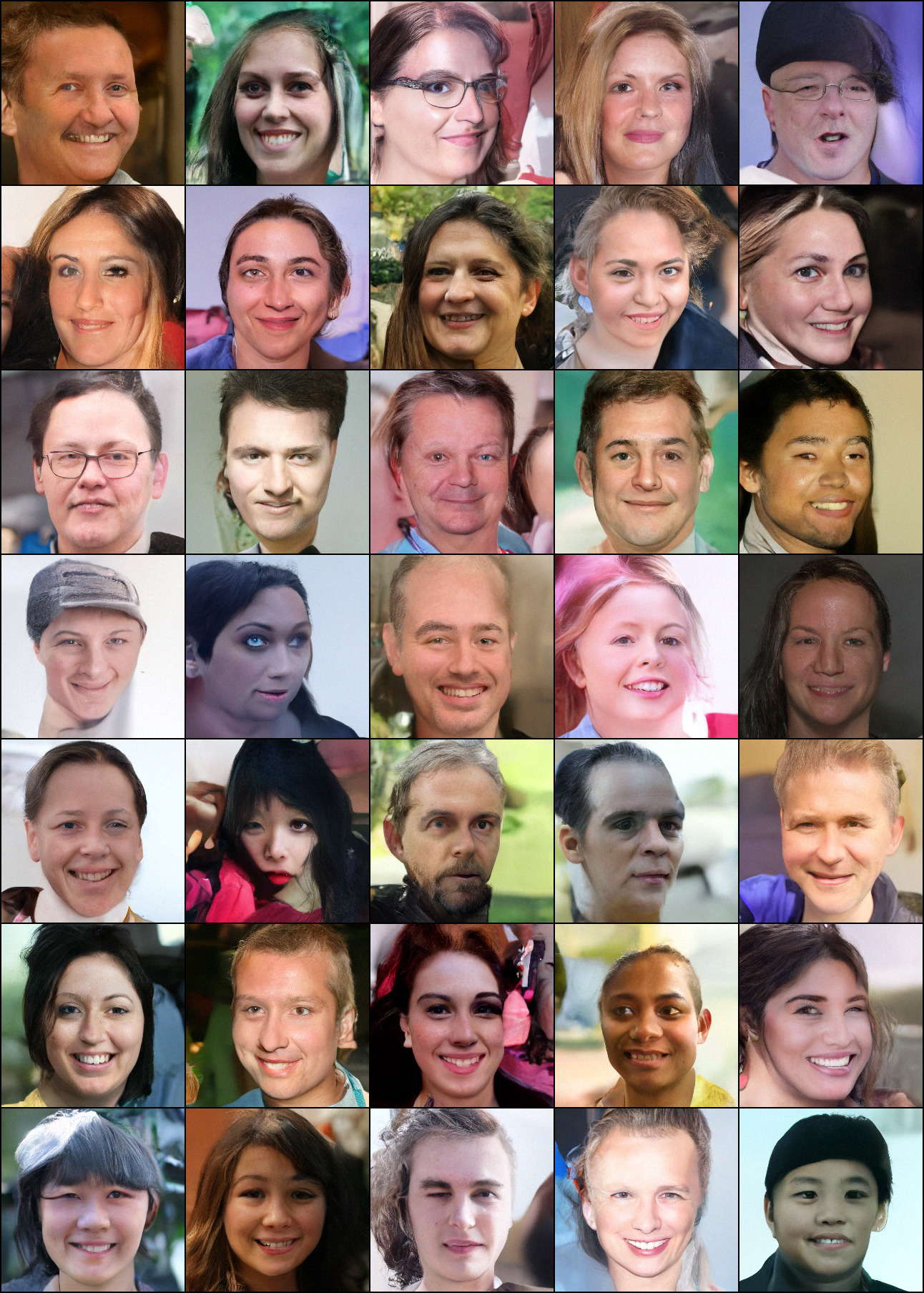}
    \caption{Uncurated FFHQ $256\times 256$ samples from NCSNv2.}
    \label{fig:ffhq_full}
\end{figure}
\vspace*{\fill}
\end{document}